\documentclass[runningheads]{llncs}
\usepackage{graphicx}

\usepackage{tikz}
\usepackage{amsmath,amssymb} 
\usepackage[accsupp]{axessibility}  

\usepackage{graphicx}
\usepackage[caption=false]{subfig}

\usepackage[utf8]{inputenc} 
\usepackage[T1]{fontenc}    
\usepackage{url}            
\usepackage{booktabs}       
\usepackage{amsfonts}       
\usepackage{nicefrac}       
\usepackage{microtype}      
\usepackage{xcolor}         
\usepackage{resizegather}
\usepackage{enumitem}
\usepackage{cancel}
\usepackage{comment}
\definecolor{pearDark}{HTML}{2980B9}
\definecolor{pearThree}{HTML}{E74C3C}
\definecolor{pearDarker}{HTML}{1D2DEC}
\usepackage{colortbl}
\usepackage{threeparttable}

\usepackage{xspace}
\makeatletter
\DeclareRobustCommand\onedot{\futurelet\@let@token\@onedot}
\def\@onedot{\ifx\@let@token.\else.\null\fi\xspace}

\def\eg{\emph{e.g}\onedot}

\makeatother

\usepackage[pagebackref,breaklinks,colorlinks]{hyperref}

\hypersetup{
	colorlinks,
	citecolor=pearDark,
	linkcolor=pearThree,
breaklinks=true,
	urlcolor=pearDarker}

\usepackage[capitalize]{cleveref}
\crefname{section}{Sec.}{Secs.}
\Crefname{section}{Section}{Sections}
\Crefname{table}{Table}{Tables}
\crefname{table}{Tab.}{Tabs.}

\begin{document}

\pagestyle{headings}
\mainmatter

\newtheorem{prop}{Proposition}
\def\E{{\rm E}\,}

\title{Decoupled Contrastive Learning}

\titlerunning{Decoupled Contrastive Learning}

\author{
Chun-Hsiao Yeh\inst{1,2} \and
Cheng-Yao Hong\inst{1} \and
Yen-Chi Hsu\inst{1,3} \and
Tyng-Luh Liu\thanks{Corresponding author. E-mail:\texttt{liutyng@iis.sinica.edu.tw}}\inst{1} \and \\
Yubei Chen\inst{4} 
\and Yann LeCun\inst{4,5}
}

\authorrunning{Yeh et al.}

\institute{IIS, Academia Sinica, Taiwan \and UC Berkeley \and
National Taiwan University \and Meta AI Research \and New York University
\\
\email{
\{sensible,yenchi,liutyng\}@iis.sinica.edu.tw}, \\
\email{daniel\_yeh@berkeley.edu}, \email{\{yubeic,yann\}@fb.com}
}

\maketitle

\begin{abstract}
Contrastive learning (CL) is one of the most successful paradigms for self-supervised learning (SSL). In a principled way, it considers two augmented ``views'' of the same image as \emph{positive} to be pulled closer, and all other images as \emph{negative} to be pushed further apart. However, behind the impressive success of CL-based techniques, their formulation often relies on heavy-computation settings, including large sample batches, extensive training epochs, etc. We are thus motivated to tackle these issues and establish a simple, efficient, yet competitive baseline of contrastive learning. Specifically, we identify, from theoretical and empirical studies, a noticeable \emph{negative-positive-coupling} (NPC) effect in the widely used InfoNCE loss, leading to unsuitable learning efficiency concerning the batch size. By removing the NPC effect, we propose decoupled contrastive learning (DCL) loss, which removes the positive term from the denominator and significantly improves the learning efficiency. DCL achieves competitive performance with less sensitivity to sub-optimal hyperparameters, requiring neither large batches in SimCLR, momentum encoding in MoCo, or large epochs. We demonstrate with various benchmarks while manifesting robustness as much less sensitive to suboptimal hyperparameters. Notably, SimCLR with DCL achieves $68.2\%$ ImageNet-1K top-1 accuracy using batch size $256$ within $200$ epochs pre-training, outperforming its SimCLR baseline by $6.4\%$. Further, DCL can be combined with the SOTA contrastive learning method, NNCLR, to achieve $72.3\%$ ImageNet-1K top-1 accuracy with $512$ batch size in $400$ epochs, which represents a new SOTA in contrastive learning. We believe DCL provides a valuable baseline for future contrastive SSL studies.


\end{abstract}

\begin{figure*}[t!]
    \centering
    \includegraphics[width=\textwidth]{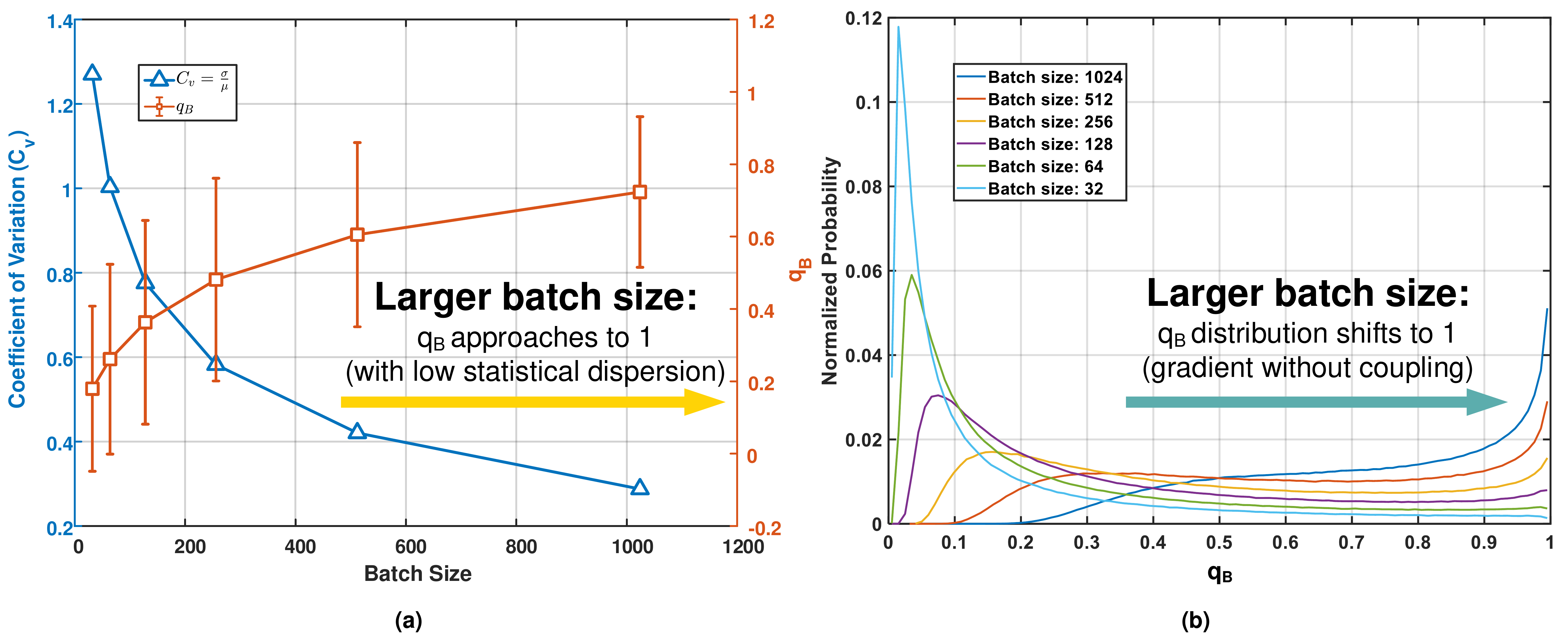}
    \caption{An overview of the batch size issue is that general contrastive approaches need large batch sizes to perform better: (a) shows the NPC multiplier $q_B$ in different batch sizes. 
    As the batch size gradually increases, the $q_B$ will approach to $1$ with a small coefficient of variation ($C_v = \sigma / \mu$); and (b) illustrates the distribution of $q_B$ with various batch sizes and indicates that the mode value of $q_B$ will shift towards $1$ when the batch size increases. Note that the $\sigma$ and $\mu$ are the standard deviation and mean of $q_B$, respectively. The coefficient of variation, $C_v$, measures the dispersion of a frequency distribution.
    } 
    \label{fig:coupling}
\end{figure*}

\section{Introduction}
As a fundamental task in machine learning, representation learning aims to extract useful information from the raw data for the downstream tasks. It has been regarded as a long-acting goal over the past decades. Recent progress on representation learning has achieved a significant milestone over self-supervised learning (SSL), facilitating feature learning with its competence in exploiting massive raw data without any annotated supervision. In the early stage of SSL, representation learning has focused on exploiting pretext tasks, which are addressed by generating pseudo-labels to the unlabeled data through different transformations, such as solving jigsaw puzzles~\cite{noroozi2016unsupervised}, colorization~\cite{zhang2016colorful} and rotation prediction~\cite{gidaris2018unsupervised}. Though these approaches succeed in computer vision, there is a large gap between these methods and supervised learning. Recently, there has been a significant advancement in using contrastive learning~\cite{wu2018unsupervised,oord2018representation,tian2019contrastive,he2020momentum,chen2020simple} for self-supervised pre-training, which significantly closes the gap between the SSL method and supervised learning. Contrastive SSL methods, e.g., SimCLR~\cite{chen2020simple}, in general, try to pull different views of the same instance close and push different instances far apart in the representation space.

Despite the evident progress of the state-of-the-art contrastive SSL methods, there have been facing several challenges into future development in this direction, including 1) The SOTA models, \eg, \cite{he2020momentum} may require specific structures such as the momentum encoder and large memory queues, which may complicate the underlying representation learning. 2) The contrastive SSL models, \eg, \cite{chen2020simple} often depend on large batch size and huge epoch numbers to achieve competitive performance, posing a computational challenge for academia to explore this direction. 3) They tend to be sensitive to hyperparameters and optimizers, introducing additional difficulty reproducing the results on various benchmarks.

Through the analysis of the widely adopted InfoNCE loss in contrastive learning, we identified a negative-positive-coupling (NPC) multiplier $q_B$ in the gradient as shown in Proposition~\ref{prop:coupling}. The NPC multiplier modulates the gradient of each sample, and it reduces the learning efficiency due to easy SSL classification tasks: 1) when a positive sample is very close to the anchor; 2) when negative samples are far away from the anchor; and 3) when there is only a small number of negative samples (i.e., a small batch size). A less-informative (nearby) positive view would reduce the gradient from a batch of informative negative samples or vice versa. Such a coupling exacerbates when smaller batch sizes are used.

Meanwhile, we also investigate the relationship between $q_B$ and batch size through the baseline, SimCLR. As can be seen in Figure~\ref{fig:coupling}, the distribution of $q_B$ has a strong positive correlation with the batch size. Figure~\ref{fig:coupling}(a) shows that when batch size gradually increases, $q_B$ not only approaches $1$ but also reduces the coefficient of variation $C_v$. The distribution with larger $C_v$ has low statistical dispersion and vice versa. Figure~\ref{fig:coupling}(b) indicates that the mode value of $q_B$ will also shift from $0$ to $1$ when the batch size becomes larger. Hence, it is reasonable to fix the value of $q_B$, alleviating the influence of batch size.

By removing the coupling term from the Info-NCE loss, we reach a new formulation, the \textit{decoupled contrastive learning} (DCL). The new objective function significantly improves the training efficiency with less sensitivity to sub-optimal hyper-parameters requires neither large batches, momentum encoding, or large epochs to achieve competitive performance on various benchmarks. The main contributions of the proposed DCL can be characterized as follows:

\begin{itemize}
  \item [1)] 
  We provide both theoretical analysis and empirical evidence to show the NPC effect in the InfoNCE-based contrastive learning;
  
  \item [2)] 
  We introduce DCL objective, which casts off the NPC coupling phenomenon, significantly improves the training efficiency, and it is less sensitive to sub-optimal hyper-parameters; 
  
  \item [3)] Extensive experiments are provided to show the effectiveness of the proposed method that DCL achieves competitive performance \textbf{without} large batch sizes, large training epochs, momentum encoding, or additional tricks such as stop-gradient and multi-cropping, etc. This leads to a plug-and-play improvement to the widely adopted InfoNCE-based contrastive learning;
  
  \item [4)] We show that DCL can be easily combined with the SOTA contrastive methods, e.g. NNCLR~\cite{dwibedi2021little}, to achieve further improvements.
\end{itemize}

\section{Related Work}

\begin{figure}[t!]
\centering
\includegraphics[width=\textwidth]{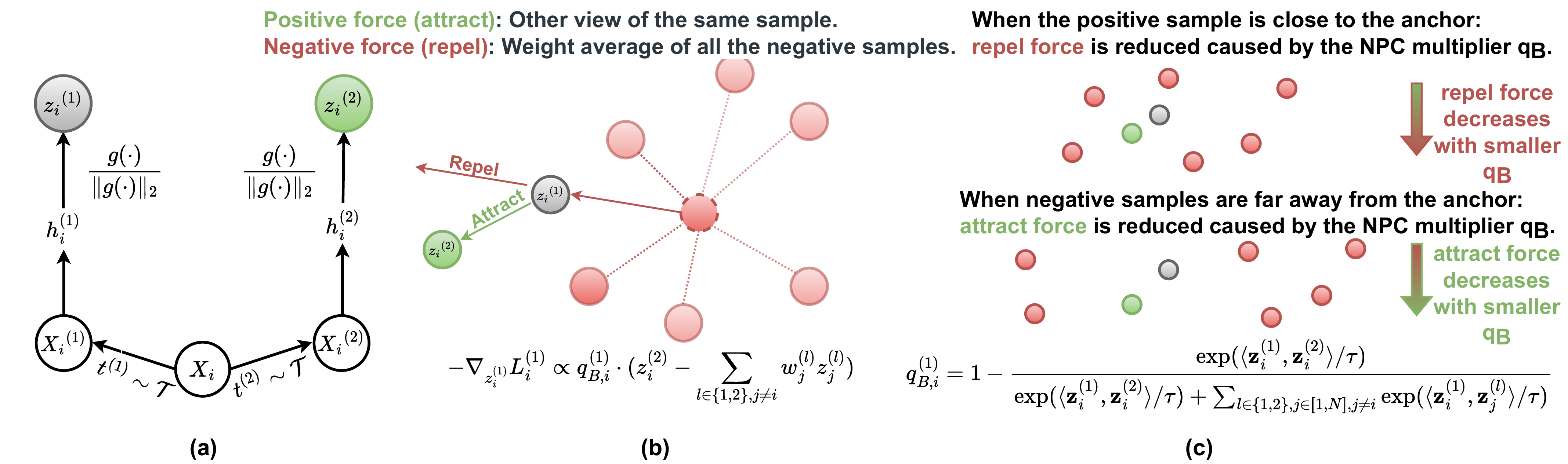}
\caption{Contrastive learning and negative-positive coupling (NPC). (a) In SimCLR, each sample $\mathbf{x}_i$ has two augmented views $\{\mathbf{x}_i^{(1)}, \mathbf{x}_i^{(2)}\}$. They are encoded by the same encoder $f$ and further projected to $\{\mathbf{z}_i^{(1)}, \mathbf{z}_i^{(2)}\}$ by a normalized MLP. (b) According to Equation~\ref{eq:NPC}. For the view $\mathbf{x}_i^{(1)}$, the cross-entropy loss $L_i^{(1)}$ leads to a positive force $\mathbf{z}_i^{(2)}$, which comes from the other view $\mathbf{x}_i^{(2)}$ of $\mathbf{x}$ and a negative force, which is a weighted average of all the negative samples, i.e. $\{\mathbf{z}_j^{(l)}|l\in\{1,2\},j\neq i\}$. However, the gradient $-\nabla_{\mathbf{z}_{i}^{(2)}}L_{i}^{(1)}$ is proportional to the NPC multiplier. (c) We show two cases when the NPC term affects learning efficiency. The positive sample is close to the anchor and less informative on the top. However, the gradient from the negative samples is also reduced. On the bottom, when the negative samples are far away and less informative, the learning rate from the positive sample is mistakenly reduced. In general, the NPC multiplier from the InfoNCE loss makes the SSL task simpler to solve, leading to reduced learning efficiency.}
\label{fig:DCL}
\end{figure}

\noindent\textbf{Contrastive Learning.}
Contrastive learning (CL) constructs positive and negative sample pairs to extract information from the data itself. In CL, each anchor image in a batch has only one positive sample to construct a positive sample pair~\cite{HadsellCL06,chen2020simple,he2020momentum}. CPC~\cite{oord2018representation} predicts the future output of sequential data by using current output as prior knowledge, which can improve the feature representing the ability of the model. Instance discrimination~\cite{wu2018unsupervised} proposes a non-parametric cross-entropy loss to optimize the model at the instance level. Inv. spread~\cite{ye2019unsupervised} makes use of data augmentation invariant and the spread-out property of instance to learn features. MoCo~\cite{he2020momentum} proposes a dictionary to maintain a negative sample set, thus increasing the number of negative sample pairs. Different from the aforementioned self-supervised CL approaches, \cite{abs-2004-11362} proposes a supervised CL that considers all the same categories as positive pairs to increase the utility of images.

\smallskip\noindent\textbf{Collapsing Issue on the Number of Negatives.}
In CL, the objective is to maximize the mutual information between the positive pairs. However, to avoid the ``\textit{collapsing output}'', vast quantities of negative samples are needed so that the learning objectives obtain the maximum similarity and have the minimum similarity with negative samples. For instance, in SimCLR~\cite{chen2020simple}, training requires many negative samples, leading to a large batch size (i.e., 4096). Furthermore, to optimize such a huge batch, a specially designed optimizer LARS~\cite{you2017large} is used. Similarly, MoCo~\cite{he2020momentum} needs a vast queue (i.e., 65536) to achieve competitive performance. BYOL~\cite{GrillSATRBDPGAP20} does not collapse output without using any negative samples by considering all the images are positive and to maximize the similarity of ``projection'' and ``prediction '' features. On the other hand, SimSiam~\cite{abs-2011-10566} leverages the Siamese network to introduce inductive biases for modeling invariance. With the small batch size (i.e., 256), SimSiam is a rival to BYOL (i.e., 4096). Unlike both approaches that achieved their success through empirical studies, this paper tackles from a theoretical perspective, proving that an intertwined multiplier $q_B $ of positive and negative is the main issue to contrastive learning.

\smallskip\noindent\textbf{Batch Size Sensitivity on InfoNCE.}
Several works of literature focus on batch size sensitivity concerning the InfoNCE objective function. \cite{tsai2021self} proposes an objective based on relative predictive coding that maintains the balance between training stability and batch size sensitivity. \cite{hjelm2018learning} follows the~\cite{belghazi2018mutual} and extends the idea between the local and global features. \cite{ozair2019wasserstein} proposes a Wasserstein distance to prevent the encoder from learning any other differences between unpaired samples. \cite{DBLP:conf/nips/KalantidisSPWL20} and~\cite{DBLP:conf/iclr/RobinsonCSJ21} learn better representation by sampling hard negatives, particularly for small batches. Other recent works~\cite{zhu2020eqco,ermolov2021whitening} aim to mitigate the issue of small batch size in InfoNCE loss. Although the basic principle of recent works and DCL is derived from InfoNCE objective function, we provide a novel perspective to support the decoupling between positive and negative terms in InfoNCE loss is essential. Simply removing the term from the denominator pre-training to positive pairs can drastically improve the performance and keep the objective function invariant to batch size sensitivity.

\section{Decouple Negative and Positive Samples in Contrastive Learning}
\label{sec:3}
We choose to start from SimCLR because of its conceptual simplicity. Given a batch of $N$ samples (e.g. images), $\{\mathbf{x}_1,\dots, \mathbf{x}_{N}\}$, let $\mathbf{x}^{(1)}_i, \mathbf{x}^{(2)}_i$ be two augmented views of the sample $x_i$ and $B$ be the set of all of the augmented views in the batch, i.e. $B=\{\mathbf{x}^{(k)}_i|k\in\{1,2\}, i\in[\![1,N]\!]\}$. As shown by Figure~\ref{fig:DCL}(a), each of the views $\mathbf{x}_i^{(k)}$ is sent into the same encoder network $f$ and the output $\mathbf{h}_i^{(k)}=f(\mathbf{x}_i^{(k)})$ is then projected by a normalized MLP projector that $\mathbf{z}_i^{(k)} = g(\mathbf{h}_i^{(k)})/\|g(\mathbf{h}_i^{(k)})\|$.  For each augmented view $\mathbf{x}_i^{(k)}$, SimCLR solves a classification problem by using the rest of the views in $B$ as targets, and assigns the only positive label to $\mathbf{x}_i^{(u)}$, where $u\neq k$. So SimCLR creates a cross-entropy loss function $L_i^{(k)}$ for each view $\mathbf{x}^{(k)}_i$, and the overall loss function is $L = \sum_{k\in\{1,2\}, i\in[\![1,N]\!]}{L_{i}^{(k)}}$.
\begin{align}
L_i^{(k)} =  -\log{\frac{\exp(\langle \mathbf{z}_i^{(1)},\mathbf{z}_i^{(2)} \rangle/\tau)}{\exp(\langle \mathbf{z}_i^{(1)},\mathbf{z}_i^{(2)} \rangle/\tau) + U_{i,k}}},
\label{eq:simclr_loss}
\end{align}\normalsize
where
\begin{align}
U_{i,k} = \sum_{l\in\{1,2\}, j\in[\![1,N]\!],j\neq i}{\exp(\langle \mathbf{z}_i^{(k)},\mathbf{z}_j^{(l)} \rangle/\tau)}    
\label{eq:negative_sum}
\end{align}
means the summation of negative terms for the view $k$ of the sample $i$.

\begin{prop}
\label{prop:coupling}{\bf :}
There exists a negative-positive coupling (NPC) multiplier $q_{B,i}^{(1)}$ in the gradient of $L_i^{(1)}$:
\begin{align}
\left\{\begin{array}{l}
    -\nabla_{\mathbf{z}_{i}^{(1)}}L_{i}^{(1)} = \\ \frac{q_{B,i}^{(1)}}{\tau}
    \left( \mathbf{z}_i^{(2)} - \sum_{l\in\{1,2\}, j\in[\![1,N]\!],j\neq i}{\frac{\exp{\langle \mathbf{z}_i^{(1)},\mathbf{z}_j^{(l)} \rangle/\tau}}{U_{i,1}}}\cdot \mathbf{z}_j^{(l)}\right) \\
    -\nabla_{\mathbf{z}_{i}^{(2)}}L_{i}^{(1)} = \frac{q_{B,i}^{(1)}}{\tau}\cdot \mathbf{z}_i^{(1)}\\
    -\nabla_{\mathbf{z}_{j}^{(l)}}L_{i}^{(1)} = - \frac{q_{B,i}^{(1)}}{\tau}\frac{\exp{\langle \mathbf{z}_i^{(1)},\mathbf{z}_j^{(l)} \rangle/\tau}}{U_{i,1}}\cdot \mathbf{z}_i^{(1)}
\end{array} \right.
\label{eq:gradient_Li}
\end{align}
where the NPC multiplier $q_{B,i}^{(1)}$ is:
\begin{align}
\label{eq:NPC}
q_{B,i}^{(1)} = 1 - \frac{\exp(\langle \mathbf{z}_i^{(1)},\mathbf{z}_i^{(2)} \rangle/\tau)}{\exp(\langle \mathbf{z}_i^{(1)},\mathbf{z}_i^{(2)}\rangle/\tau) + U_{i,1}}
\end{align}
and $U_{i,1} = {\sum_{l\in\{1,2\}, j\in[\![1,N]\!], j\neq i}{\exp(\langle \mathbf{z}_i^{(1)},\mathbf{z}_j^{(l)} \rangle/\tau)}}$. Due to the symmetry, a similar NPC multiplier $q_{B,i}^{(k)}$ exists in the gradient of $L_i^{(k)}, k\in\{1,2\}, i\in[\![1,N]\!]$.
\end{prop}

As we can see, all of the partial gradients in Equation~\ref{eq:gradient_Li} are modified by the common NPC multiplier $q_{B,i}^{(k)}$ in Equation~\ref{eq:NPC}. Equation~\ref{eq:NPC} makes intuitive sense: when the SSL classification task is easy, the gradient would be reduced by the NPC term. However, the positive samples and negative samples are strongly coupled. When the negative samples are far away and less informative (easy negatives), the gradient from an informative, positive sample would be reduced by the NPC multiplier $q_{B,i}^{(1)}$. On the other hand, when the positive sample is close (easy positive) and less informative, the gradient from a batch of informative negative samples would also be reduced by the NPC multiplier. When the batch size is smaller, the SSL classification problem can be significantly simpler to solve. As a result, the learning efficiency can be significantly reduced with a small batch size setting.

Figure~\ref{fig:coupling}(b) shows the NPC multiplier $q_{B}$ distribution shift w.r.t. different batch sizes for a pre-trained SimCLR baseline model. While all of the shown distributions have prominent fluctuation, the smaller batch size makes $q_{B}$ cluster towards $0$, while the larger batch size pushes the distribution towards $\delta(1)$. Figure~\ref{fig:coupling}(a) shows the averaged NPC multiplier $\langle q_{B} \rangle$ changes w.r.t. the batch size and the relative fluctuation. The small batch sizes introduce significant NPC fluctuation. Based on this observation, we propose to remove the NPC multipliers from the gradients, which corresponds to the case $q_{B, N \to \infty}$. This leads to the decoupled contrastive learning formulation. \cite{wang2020understanding} also proposes an alignment \& uniformity loss which does not have the NPC. However, a similar analysis introduces negative-negative coupling from different positive samples. In other words, \cite{wang2020understanding} considers all the negative samples in the batch together, which may cause the gradient to be dominated by a specific negative pair. In Appendix~5, we provide a thorough discussion and demonstrate the advantage of DCL loss against~\cite{wang2020understanding}.

\begin{prop}
\label{prop:decoupling} {\bf the DCL Loss:}
Removing the positive pair from the denominator of Equation~\ref{eq:simclr_loss} leads to a decoupled contrastive learning loss. If we remove the NPC multiplier $q_{B,i}^{(k)}$ from Equation~\ref{eq:gradient_Li}, we reach a decoupled contrastive learning loss $L_{DC} = \sum_{k\in\{1,2\}, i\in[\![1,N]\!]}{L_{DC,i}^{(k)}}$, where $L_{DC,i}^{(k)}$ is:
\begin{align}
L_{DC,i}^{(k)} &= -\log{\frac{\exp(\langle \mathbf{z}_i^{(1)},\mathbf{z}_i^{(2)} \rangle/\tau)}{\bcancel{\exp(\langle \mathbf{z}_i^{(1)},\mathbf{z}_i^{(2)} \rangle/\tau)} + U_{i,k}}} \\
&= -\langle \mathbf{z}_i^{(1)},\mathbf{z}_i^{(2)} \rangle/\tau + \log{U_{i,k}}
\label{eq:DC}
\end{align}
\end{prop}
The proofs of Proposition~\ref{prop:coupling} and \ref{prop:decoupling} are given in Appendix. Further, we can generalize the loss function $L_{DC}$ to $L_{DCW}$ by introducing a weighting function for the positive pairs i.e. $L_{DCW} = \sum_{k\in\{1,2\}, i\in[\![1,N]\!]}{L_{DCW,i}^{(i,k)}}$.
\begin{align}
L_{DCW,i}^{(k)} = -w(\mathbf{z}_i^{(1)},\mathbf{z}_i^{(2)})(\langle \mathbf{z}_i^{(1)},\mathbf{z}_i^{(2)} \rangle/\tau) + \log{U_{i,k}}
\label{eq:DCW}
\end{align}
where we can intuitively choose $w$ to be a negative von Mises-Fisher weighting function that $w(\mathbf{z}_i^{(1)},\mathbf{z}_i^{(2)}) = 2 - \frac{\exp(\langle \mathbf{z}_i^{(1)},\mathbf{z}_i^{(2)} \rangle/\sigma)}{\E_i \left[\exp(\langle \mathbf{z}_i^{(1)},\mathbf{z}_i^{(2)} \rangle/\sigma)\right]}$ and $\E \left[w\right] = 1$. $L_{DC}$ is a special case of $L_{DCW}$ and we can see that $\lim_{\sigma\to \infty}{L_{DCW}} = L_{DC}$. The intuition behind $w(\mathbf{z}_i^{(1)},\mathbf{z}_i^{(2)})$ is that there is more learning signal when a positive pair of samples are far from each other, and $E\left[w(\mathbf{z}_i^{(1)},\mathbf{z}_i^{(2)})\langle \mathbf{z}_i^{(1)},\mathbf{z}_i^{(2)} \rangle\right] \approx E\left[\langle \mathbf{z}_i^{(1)},\mathbf{z}_i^{(2)} \rangle\right]$. Other similar weight functions also provide similar results. In general, we find such a weighting function, which gives a larger weight to the hard positives tend to increase the representation quality.

\section{Experiments}
\label{others}

This section empirically evaluates the proposed decoupled contrastive learning (DCL) and compares it to general contrastive learning methods. We summarize the experiments and analysis as the following: (1) the proposed work significantly outperforms the general InfoNCE-based contrastive learning on both large-scale and small-scale vision benchmarks; (2) we show that the enhanced version of DCL, DCLW, could further improve the representation quality; and (3) we further analyze DCL with ablation studies on ImageNet-1K, hyperparameters, and few learning epochs, which shows fast convergence of the proposed DCL. Note that all the experiments are conducted with 8 Nvidia V100 GPUs on a single machine.

\subsection{Implementation Details}
\label{sub:4.1}

\smallskip \noindent \textbf{ImageNet.}
For a fair comparison on ImageNet data, we implement the proposed decoupled structure, DCL, by following SimCLR~\cite{chen2020simple} with ResNet-50~\cite{he2016deep} as the encoder backbone and use cosine annealing schedule with SGD optimizer. We set the temperature $\tau$ to 0.1 and the latent vector dimension to 128. Following the OpenSelfSup benchmark~\cite{openselfsup}, we evaluate the pre-trained models by training a linear classifier with frozen learned embedding on ImageNet data. We further consider evaluating DCL on ImageNet-100, a selected subset of 100 classes of ImageNet-1K. Note that all models on ImageNet are trained for 200 epochs.

\smallskip \noindent \textbf{CIFAR and STL10.}
For CIFAR10, CIFAR100, and STL10, ResNet-18~\cite{he2016deep} is used as the encoder architecture. Following the small-scale benchmark~\cite{wang2020unsupervised}, we set the temperature $\tau$ to 0.07. All models are trained for 200 epochs with SGD optimizer, a base $lr = 0.03*batchsize/256$, and evaluated by k nearest neighbor (kNN) classifier. Note that on STL10, we include both the $train$ and $unlabeled$ set for model pre-training. We further use ResNet-50 as a stronger backbone by following the implementation~\cite{pyTorchimplementation}, using the same backbone and hyperparameters.

\subsection{Experiments and Analysis}

\begin{figure*}[t!]
\begin{center}
\includegraphics[width=\textwidth]{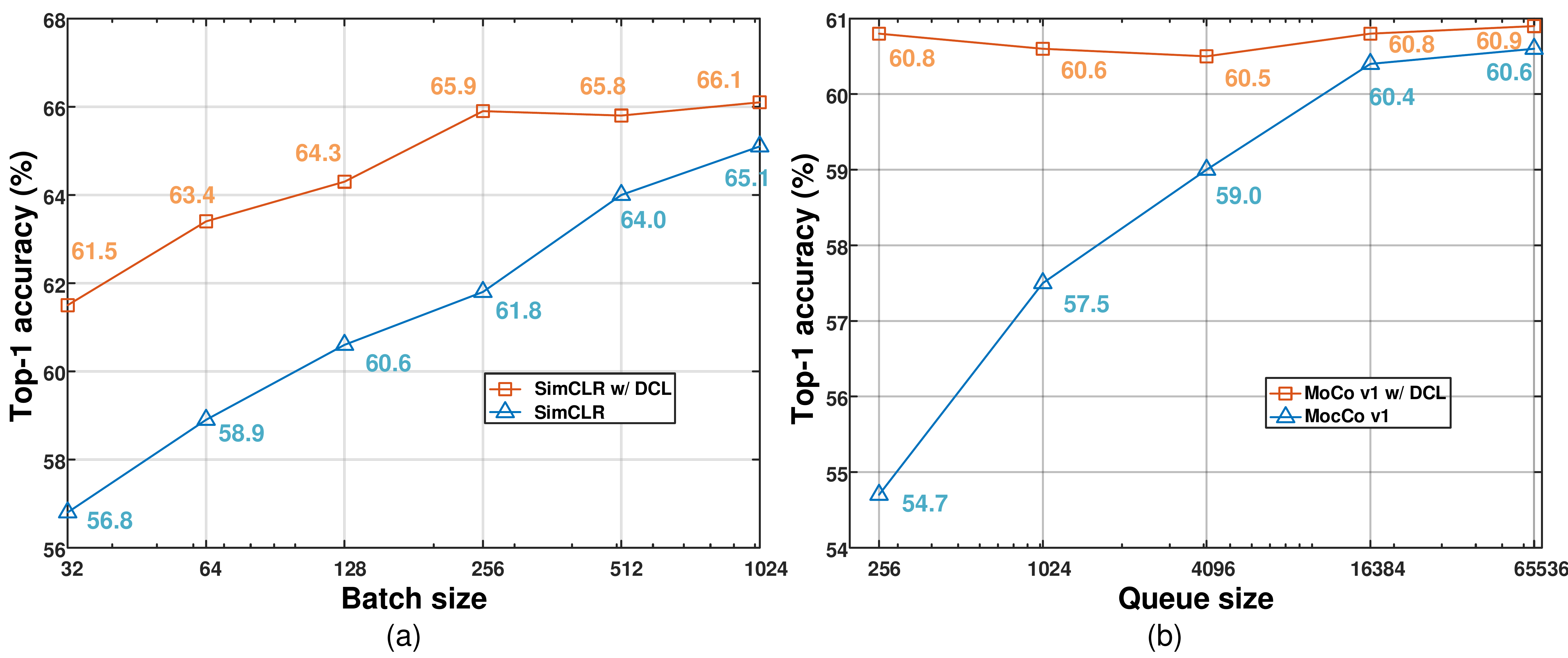}
\caption{Comparisons on ImageNet-1K with/without DCL under different numbers of (a): batch sizes for SimCLR and (b): queues for MoCo. Without DCL, the top-1 accuracy significantly drops when batch size (SimCLR) or queues (MoCo) becomes very small. Note that the temperature $\tau$ is $0.1$ for SimCLR and $0.07$ for MoCo in the comparison.}
\label{fig:mocosimclr}
\end{center}
\end{figure*}

\smallskip \noindent \textbf{DCL on ImageNet.}
This section illustrates the effect of DCL against InfoNCE-based approaches under different batch sizes and queues. The initial setup is to have 1024 batch size (SimCLR) and 65536 queues (MoCo~\cite{he2020momentum}) and gradually reduce the batch size (SimCLR) and queue (MoCo) to show the corresponding top-1 accuracy by linear evaluation. Figure~\ref{fig:mocosimclr} indicates that without DCL, the top-1 accuracy drastically drops when batch size (SimCLR) or queue (MoCo) becomes very small. While with DCL, the performance keeps steadier than baselines (SimCLR: $-4.1\%$ vs. $-8.3\%$, MoCo: $-0.4\%$ vs. $-5.9\%$).

\begin{table}[t!]
\centering
\caption{Comparisons with/without DCL under different batch sizes from 32 to 512. Results show the effectiveness of DCL on five widely used benchmarks. The performance of DCL keeps steadier than the SimCLR baseline while the batch size is varied.}
\label{tab:batchsize}
\resizebox{0.89\textwidth}{!}{
\begin{tabular}{l|ccccc}
\toprule
\multicolumn{1}{c|}{Batch Size}  &  32 & 64 & 128 & 256 & 512  \\ 
\midrule
\midrule
\multicolumn{1}{c|}{Dataset} & \multicolumn{5}{c}{ImageNet-1K (kNN / Linear)} \\ 
\midrule
Baseline (ResNet-50)        & 40.2/56.8	& 42.9/58.9 & 45.1/60.6	& 46.3/61.8 & 49.4/64.0\\ 
w/ DCL (ResNet-50)          & \cellcolor{pearDark!20}\textbf{43.7/61.5}	& \cellcolor{pearDark!20}\textbf{46.3/63.4} & \cellcolor{pearDark!20}\textbf{48.5/64.3}	& \cellcolor{pearDark!20}\textbf{49.8/65.9} & \cellcolor{pearDark!20}\textbf{50.1/65.8}\\ 
\midrule
\multicolumn{1}{c|}{Dataset} & \multicolumn{5}{c}{ImageNet-100 (kNN / Linear)} \\ 
\midrule
Baseline (ResNet-50)        & 67.8/74.2	& 71.9/77.6 & 73.2/79.3 & 74.6/80.7 & 75.4/81.3 \\ 
w/ DCL (ResNet-50)          & \cellcolor{pearDark!20}\textbf{74.9/80.8}	& \cellcolor{pearDark!20}\textbf{76.3/82.0} & \cellcolor{pearDark!20}\textbf{76.5/81.9}	& \cellcolor{pearDark!20}\textbf{76.9/83.1} & \cellcolor{pearDark!20}\textbf{76.8/82.8}\\ 
\midrule
\multicolumn{1}{c|}{Dataset} & \multicolumn{5}{c}{CIFAR10 (kNN / Linear)} \\ 
\midrule
Baseline (ResNet-18)        & 78.9/79.8	& 80.4/81.3 &	81.1/82.8	& 81.4/83.0 & 81.3/83.3\\ 
w/ DCL (ResNet-18)          & \cellcolor{pearDark!20}\bf{83.7}/\bf{85.1}	& \cellcolor{pearDark!20}\textbf{84.4/85.9} & \cellcolor{pearDark!20}\textbf{84.4/85.7} & \cellcolor{pearDark!20}\textbf{84.2/85.3} &	\cellcolor{pearDark!20}\textbf{83.5/84.7} \\
\midrule
\multicolumn{1}{c|}{Dataset} & \multicolumn{5}{c}{CIFAR100 (kNN / Linear)} \\ \hline
Baseline (ResNet-18)        & 49.4/51.3	& 50.3/53.8 &	51.8/55.3	& 52.0/56.3 &	52.4/56.8\\ 
w/ DCL (ResNet-18)          & \cellcolor{pearDark!20}\textbf{51.1/55.4}	& \cellcolor{pearDark!20}\textbf{54.3/58.3} & \cellcolor{pearDark!20}\textbf{54.6/58.9}	& \cellcolor{pearDark!20}\textbf{54.9/58.5} & \cellcolor{pearDark!20}\textbf{55.0/58.4} \\ 
\midrule
\multicolumn{1}{c|}{Dataset} & \multicolumn{5}{c}{STL10 (kNN / Linear)} \\ \hline
Baseline (ResNet-18)        & 74.1/76.2 & 77.6/77.8 & 79.3/80.0 & 80.7/81.3 & 81.3/81.5\\ 
w/ DCL (ResNet-18)          & \cellcolor{pearDark!20}\textbf{82.0/85.2}	& \cellcolor{pearDark!20}\textbf{82.8/86.3} & \cellcolor{pearDark!20}\textbf{81.8/86.1}	& \cellcolor{pearDark!20}\textbf{81.2/85.7} & \cellcolor{pearDark!20}\textbf{81.0/85.6} \\ 
\bottomrule
\end{tabular}
}
\end{table}

Specifically, Figure~\ref{fig:mocosimclr} further shows that in SimCLR, the performance with DCL improves from $61.8\%$ to $65.9\%$ under 256 batch size; MoCo with DCL improves from $54.7\%$ to $60.8\%$ under 256 queues. The comparison fully demonstrates the necessity of DCL, especially when the number of negatives is small. Although batch size increases to 1024, DCL ($66.1\%$) still improves over the SimCLR baseline ($65.1\%$).

We further observe the same phenomenon on ImageNet-100 data. Table~\ref{tab:batchsize} shows that, with DCL, the top-1 linear performance only drops $2.3\%$ compared to the InfoNCE baseline (SimCLR) of $7.1\%$ when the batch size is varied.

In summary, it is worth noting that, while the batch size is small, the strength of $q_{B, i}$, which is used to push the negative samples away from the positive sample, is also relatively weak. This phenomenon tends to reduce the efficiency of learning representation. While taking advantage of DCL alleviates the performance gap between small and large batch sizes. Hence, through the analysis, we find out DCL can simply tackle the batch size issue in contrastive learning. With this considerable advantage given by DCL, general SSL approaches can be implemented with fewer computational resources or lower standard platforms. Compared to InfoNCE, DCL is more applicable across all large-scale SSL applications.

\smallskip \noindent \textbf{DCL on CIFAR and STL10.}
\label{par:dcl}
For STL10, CIFAR10, and CIFAR100, we implement DCL with ResNet-18 as encoder backbone. In Table~\ref{tab:batchsize}, it is observed that DCL also demonstrates its strong effectiveness on small-scale benchmarks. In the evaluation (kNN / Linear) summary, DCL outperforms its baseline by $4.8\%$ / $5.3\%$ (CIFAR10) and $1.7\%$ / $4.4\%$ (CIFAR100) under a small batch size 32. The accuracy (kNN / Linear) of the SimCLR baseline on STL10 is also improved significantly by $7.9\%$ / $9.0\%$.

 \begin{table}[t!]
     \centering
      \caption{Comparisons between SimCLR baseline, DCL, and DCLW. The linear and kNN top-1 ($\%$) results indicate that DCL improves baseline performance, and DCLW further provides an extra boost. Note that results are under batch size 256 and epoch 200. All models are both trained and evaluated with the same experimental settings. The backbones are ResNet-18 and ResNet-50 for CIFAR and ImageNet, respectively.}
     \label{tab:DCL-DCLW}
     \resizebox{0.8\textwidth}{!}{
     \begin{tabular}{l|cccc}
         \toprule
          Dataset      & CIFAR10 (kNN) & CIFAR100 (kNN) &ImageNet-100 (linear) & ImageNet-1K (linear)       \\
          \midrule
          \midrule
          SimCLR     & 81.4     & 52.0 & 80.7 & 61.8  \\
          DCL          & 84.2 (\textcolor{pearDark}{\bf+2.8)} & 54.9 (\textcolor{pearDark}{\bf+2.9)}  & 83.1 (\textcolor{pearDark}{\bf+2.4)} & 65.9 (\textcolor{pearDark}{\bf+4.1)}  \\
          DCLW         & \bf{84.8} (\textcolor{pearDark}{\bf+3.4)}     & \bf{55.2} (\textcolor{pearDark}{\bf+3.2)} & \bf{84.2} (\textcolor{pearDark}{\bf+3.5)} & \bf{66.9} (\textcolor{pearDark}{\bf+5.1)}  \\
         \bottomrule
     \end{tabular}
     }
\end{table}

\begin{table}[t!]
\centering
\caption{Improve the DCL model performance on ImageNet-1K with tuned hyperparameters: temperature and learning rate, and stronger image augmentation. Note that models are trained with 256 batch size and 200 epochs.}
\label{tab:strong-aug}

\begin{tabular}{l|c}
         \toprule
          ImageNet-1K (256 Batch size; 200 epoch) & Linear Top-1 Accuracy (\%)       \\
          \midrule
          \midrule
          DCL &	65.9               \\
          + optimal ($\tau, l_r$) = (0.2, 0.07)	& 67.8 (\textcolor{pearDark}{\bf+1.9})             \\
          + asymmetric augmentation~\cite{GrillSATRBDPGAP20}	            & 68.2  (\textcolor{pearDark}{\bf+0.4}) \\
         \bottomrule
     \end{tabular}

\end{table}

\smallskip \noindent \textbf{Decoupled Objective with Re-Weighting DCLW.} 
We only replace $L_{DC}$ with $L_{DCW}$ with no possible advantage from additional tricks. Both DCL and the baselines apply the same training instruction of the OpenSelfSup benchmark for fairness. Note that we empirically choose $\sigma = 0.5$ in the experiments.  
Results in Table~\ref{tab:DCL-DCLW} indicates that, DCLW achieves extra $5.1\%$ (ImageNet-1K), $3.5\%$ (ImageNet-100) gains compared to the baseline. For CIFAR data, an extra $3.4\%$ (CIFAR10) $3.2\%$ is gained from the addition of DCLW. It is worth noting that, trained with 200 epochs, DCLW reaches $66.9\%$ with batch size 256, surpassing the SimCLR baseline: $66.2\%$ with batch size 8192.

\subsection{Ablations}
\label{sub:4.4}
We perform extensive ablations on the hyperparameters of DCL on both ImageNet data and other small-scale data, i.e., CIFAR and STL10. By seeking better configurations empirically, we see that DCL gives consistent gains over the standard InfoNCE baselines (SimCLR and MoCo-v2). In other ablations, we see that DCL achieves more gains over both SimCLR and MoCo-v2, i.e., InfoNCE-based baselines, also when training for 100 epochs only. 

\smallskip \noindent \textbf{DCL Ablations on ImageNet.}
In Table~\ref{tab:strong-aug}, we have slightly improved the DCL model performance on ImageNet-1K: 1) tuned hyperparameters, temperature $\tau$ and learning rate ; 2) asymmetric image augmentation (e.g., BYOL). To obtain a stronger baseline, we conduct an empirical hyperparameter search with batch size 256 and 200 epochs. This improves DCL from 65.9\% to 67.8\% top-1 accuracy on ImageNet-1K. We further adopt the asymmetric augmentation policy from BYOL and improve DCL from 67.8\% to 68.2\% top-1 accuracy on ImageNet-1K.

\smallskip \noindent \textbf{DCL Ablations on CIFAR.}
Further experiments are conducted based on the ResNet-50 backbone and large learning epochs (i.e., 500 epochs). The DCL model with kNN eval, batch size 32, and 500 epochs of training could reach 86.1\% compared to 82.2\%. For the following experiments in Table~\ref{tab:resnet50cifar}, we show DCL ResNet-50 performance on CIFAR10 and CIFAR100. In these comparisons, we vary the batch size to show the effectiveness of DCL.

\smallskip \noindent \textbf{MoCo-v2 with DCL.}
We are aware that it is more convincing to compare the proposed DCL against a more compelling version, MoCo-v2. Comparisons on both ImageNet-1K and ImageNet-100 in Table~\ref{tab:mocov2} indicate that DCL becomes significantly more effective than MoCo-v2 when the queue size gets smaller.

\begin{table}[t]
\centering
\caption{The comparisons with/without DCL under various batch sizes from 32 to 512 on ResNet-50.}
\label{tab:resnet50cifar}
\begin{tabular}{l|ccccc|ccccc}
\toprule
Architecture@epoch &  \multicolumn{10}{c}{ResNet-50@500 epoch} \\
\midrule
\midrule
        Dataset & \multicolumn{5}{c}{CIFAR10 (kNN)} & \multicolumn{5}{|c}{CIFAR100 (kNN)} \\
        \midrule
        Batch Size  &  32 & 64 & 128 & 256 & 512 &  32 & 64 & 128 & 256 & 512   \\ 
        \midrule
        SimCLR   & 82.2	& 85.9 & 88.5 & 88.9 & 89.1 &  49.8 & 55.3 & 59.9 & 60.6 &61.1 \\
        SimCLR w/ DCL  & \cellcolor{pearDark!20}\bf{86.1} & 
        \cellcolor{pearDark!20}\bf{88.3} & \cellcolor{pearDark!20}\bf{89.9} & 
        \cellcolor{pearDark!20}\bf{90.1} & \cellcolor{pearDark!20}\bf{90.3} & \cellcolor{pearDark!20}\bf{54.3} & 
        \cellcolor{pearDark!20}\bf{58.4} & \cellcolor{pearDark!20}\bf{61.6} & 
        \cellcolor{pearDark!20}\bf{62.0} & \cellcolor{pearDark!20}\bf{62.2} \\
    \bottomrule
\end{tabular}
\end{table}

\begin{table}[t!]
\centering
\caption{Linear top-1 accuracy ($\%$) comparison with MoCo-V2 on ImageNet-1K and ImageNet-100.}
\label{tab:mocov2}
\begin{tabular}{l|ccccc|ccc}
\toprule
Queue Size  &  32 & 64 & 128 & 256 & 8192 & 64 & 256 & 65536 \\ 
\midrule
Dataset &\multicolumn{5}{c}{ImageNet-100 (Linear)} & \multicolumn{3}{|c}{ImageNet-1K (Linear)} \\
\midrule
\midrule
MoCo-v2 Baseline (ResNet-50)
& 73.7 & 76.4 & 78.7 & 78.7& 79.8 & 63.9 & 67.1 & 67.5 \\ 
MoCo-v2 w/DCL (ResNet-50) & 
\cellcolor{pearDark!20}\textbf{76.2} & 
\cellcolor{pearDark!20}\textbf{78.3} &
\cellcolor{pearDark!20}\textbf{79.6} &
\cellcolor{pearDark!20}\textbf{79.6} &
\cellcolor{pearDark!20}\textbf{80.5} &
\cellcolor{pearDark!20}\textbf{65.8} &
\cellcolor{pearDark!20}\textbf{67.6} & \cellcolor{pearDark!20}\textbf{67.7}\\ 
\bottomrule
\end{tabular}
\end{table}

\begin{table}[t]
 \centering
  \caption{ImageNet-1K top-1 accuracy (\%) on SimCLR and MoCo-v2 with/without DCL under few training epochs. We further list results under 200 epochs for clear comparison. With DCL, the performance of SimCLR trained under 100 epochs nearly reaches its performance under 200 epochs. The MoCo-v2 with DCL also reaches higher accuracy than the baseline under 100 epochs.}
 \label{tab:shortepoch}

 \begin{tabular}{l|cccc}
     \toprule
                   & SimCLR & SimCLR w/ DCL & MoCo-v2 & MoCo-v2 w/ DCL       \\
      \midrule
      \midrule
      100 Epoch    & 57.5     & \cellcolor{pearDark!20}64.6       & 63.6   & \cellcolor{pearDark!20}64.4  \\
      200 Epoch    & 61.8     & \cellcolor{pearDark!20}65.9       & 67.5   & \cellcolor{pearDark!20}67.7  \\
     \bottomrule
 \end{tabular}

\end{table}

\begin{figure}[ht!]
\begin{minipage}[t]{0.29\textwidth}
    \centering
    \subfloat[CIFAR10]{\includegraphics[width=\textwidth]{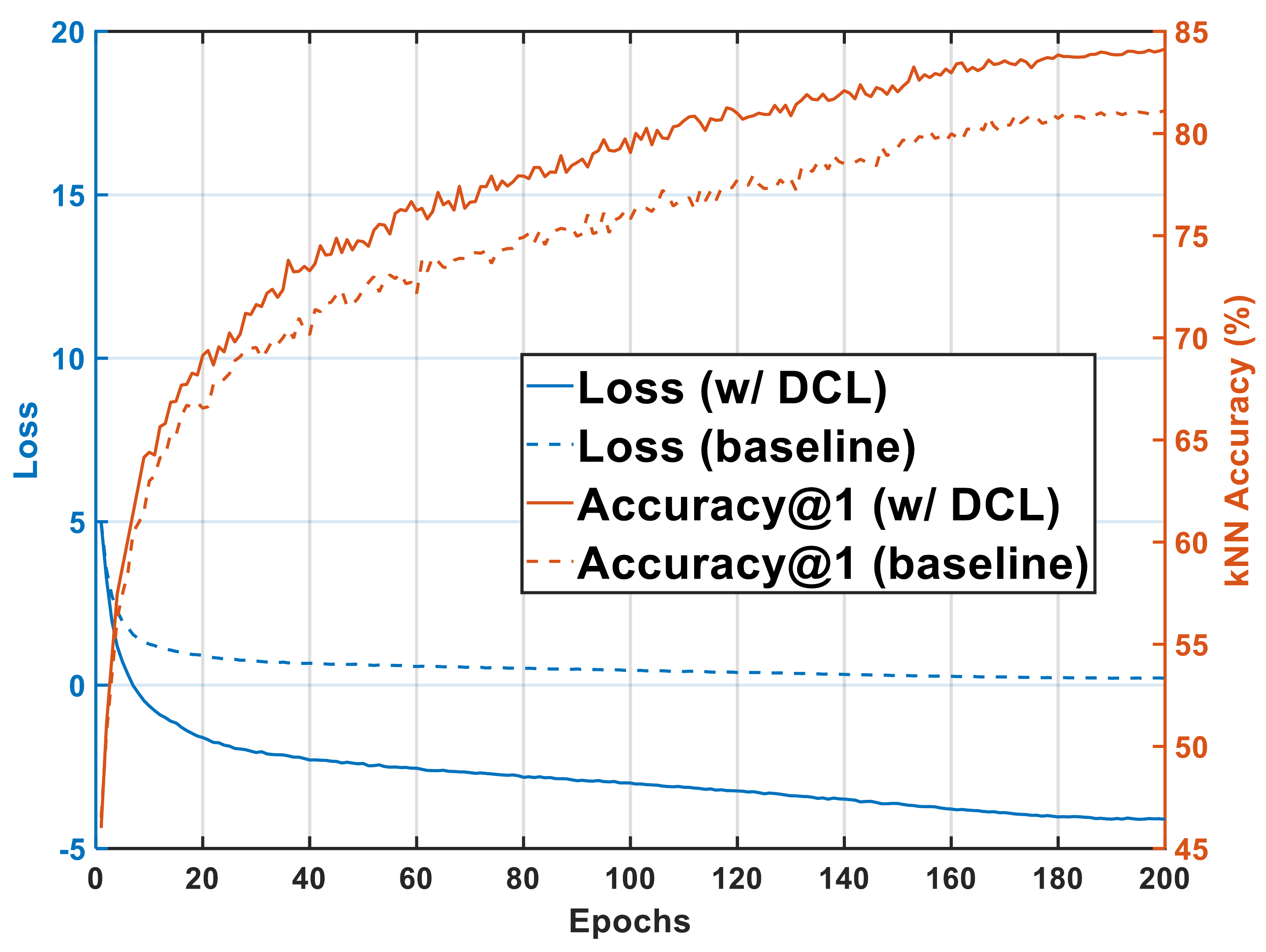}} \\
    \subfloat[STL10]{\includegraphics[width=\textwidth]{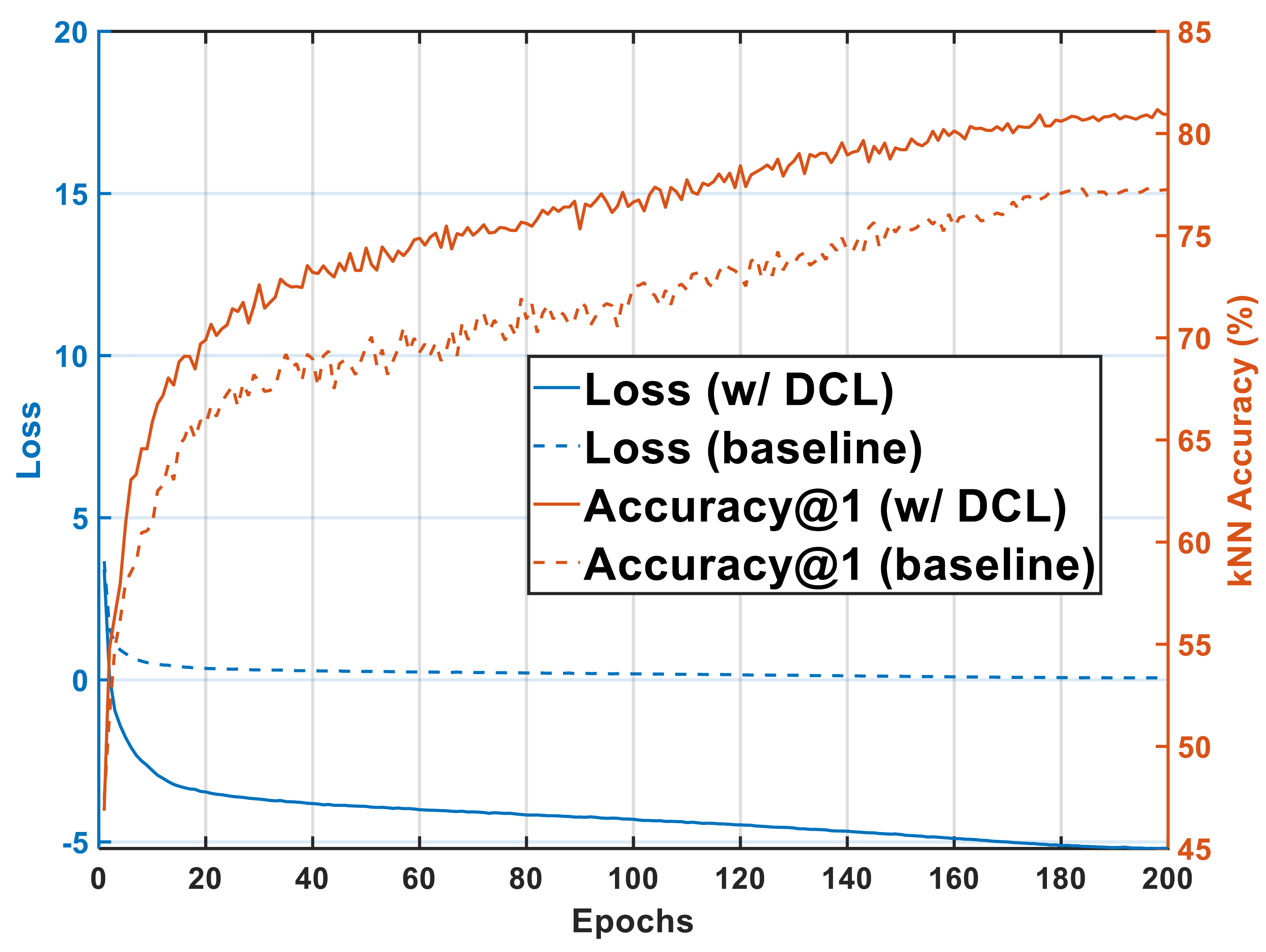}}
\end{minipage}
\hfill
\begin{minipage}[t]{0.69\textwidth}
    \centering
    \subfloat[t-SNE visualization]{\includegraphics[width=\textwidth]{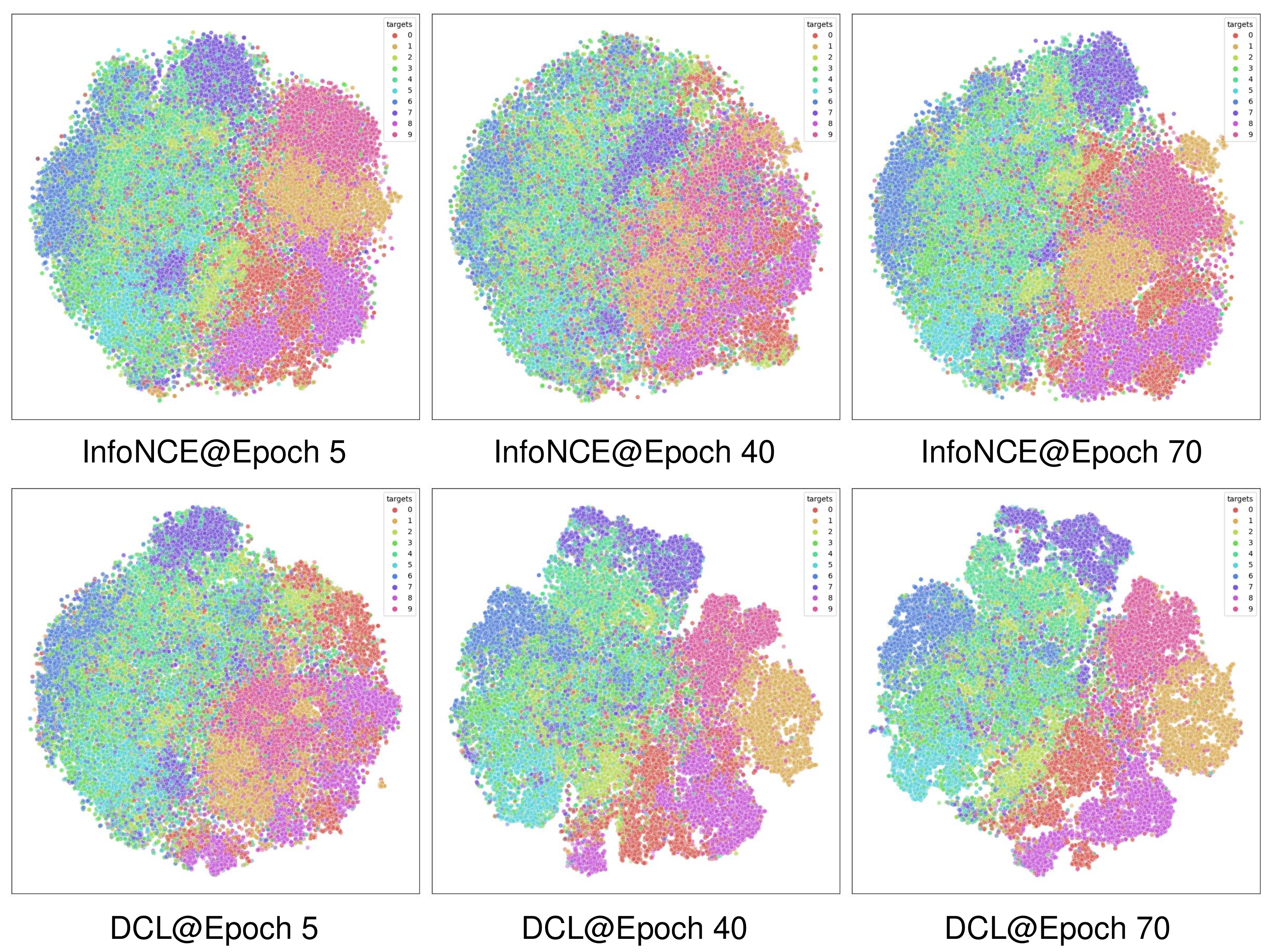}}
\end{minipage}
    \caption{Comparisons between DCL and InfoNCE-based baseline (SimCLR) on (a) CIFAR10 and (b) STL10 data. DCL speeds up the model convergence during the SSL pre-training and provides better performance than the baseline on CIFAR and STL10 data. (c) t-SNE visualization of CIFAR10 with 32 batch size. DCL shows a stronger separation force between the features than SimCLR.
    }
    \label{fig:fastconverge}
\end{figure}

\smallskip \noindent \textbf{Few Learning Epochs.} 
DCL can alleviate the shortcoming of the traditional contrastive learning framework, which needs a large batch size long learning epochs to achieve higher performance. 
The previous state-of-the-art, SimCLR, heavily relies on large quantities of learning epochs to obtain high top-1 accuracy. (e.g., $69.3\%$ with up to 1000 epochs). DCL aims to achieve higher learning efficiency with few learning epochs. We demonstrate the effectiveness of DCL in InfoNCE-based frameworks SimCLR and MoCo-v2~\cite{chen2020improved}. We choose the batch size of 256 (queue of 65536) as the baseline and train the model with only 100 epochs. We make sure other parameter settings are the same for a fair comparison. Table~\ref{tab:shortepoch} shows the result on ImageNet-1K using linear evaluation. With DCL, SimCLR can achieve $64.6\%$ top-1 accuracy with only 100 epochs compared to SimCLR baseline: $57.5\%$; MoCo-v2 with DCL reaches $64.4\%$ compared to MoCo-v2 baseline: $63.6\%$ with 100 epochs pre-training.

We further demonstrate that, with DCL, learning representation becomes faster during the early stage of training compared to the InfoNCE-based learning scheme. The reason is that DCL successfully solves the decoupled issue between positive and negative pairs. Figure~\ref{fig:fastconverge} on (a) CIFAR10 and (b) STL10 shows that DCL improves the speed of convergence and reaches higher performance than the baseline on CIFAR and STL10 data. The t-SNE visualization in Figure~\ref{fig:fastconverge} (c) also supports the proposed theoretical derivation that removing the batch-size dependent impact (i.e., NPC multiplier) should improve representation learning abilities over the InfoNCE-based learning scheme.

\section{Discussion}

\smallskip \noindent \textbf{Comparison with other SOTA SSL Approaches.}
The primary goal of this work is to provide an efficient and effective improvement to the widely used InfoNCE-based contrastive learning, where we decouple the positive and negative terms to achieve better representation quality. DCL is less sensitive to suboptimal hyperparameters and achieves competitive results with minimal requirements. Its effectiveness does not rely on large batch size and learning epochs, momentum encoding, negative sample queues, or additional tactics (e.g., stop-gradient and multi-cropping). Overall, DCL provides a more robust baseline for the contrastive-based SSL approaches. Though this work aims not to provide a SOTA SSL approach, DCL can be combined with the SOTA contrastive learning methods, such as NNCLR~\cite{dwibedi2021little}, to achieve better performance without large batch size and learning epochs. In Table~\ref{tab:sotassl-imagenet1k}, we provide extensive comparisons to SOTA SSL approaches on ImageNet-1K to validate the effectiveness of DCL. In Table~\ref{tab:compare-cifar-img100}, we further show that DCL achieves competitive results compared to VICReg~\cite{bardes2021vicreg}, Barlow Twins~\cite{zbontar2021barlow}, SimSiam~\cite{abs-2011-10566}, SwAV~\cite{caron2018deep}, and DINO~\cite{abs-2104-14294} on ImageNet-100 and CIFAR10.

\begin{table}[t!]
\centering
\caption{Linear top-1 accuracy ($\%$) comparison of SSL approaches on ImageNet-1K. Given lower computational budget, DCL model are better than recent SOTA approaches. Its effectiveness \textbf{does not rely on} large batch size and epochs (SimCLR~\cite{chen2020simple}, NNCLR~\cite{dwibedi2021little}), momentum encoding (BYOL~\cite{GrillSATRBDPGAP20}, MoCo-v2~\cite{chen2020improved}), or other tricks such as stop-gradient (SimSiam~\cite{abs-2011-10566}) and multi-cropping (SwAV~\cite{caron2020unsupervised}).}
\label{tab:sotassl-imagenet1k}
\resizebox{\textwidth}{!}{
\begin{tabular}{l|ccc|cc|c|c|c}
\toprule
ResNet-50 w/ & SimCLR & BYOL & SwAV & MoCo-v2 & SimSiam & Barlow Twins & NNCLR & NNCLR +DCL \\ 
\midrule
\midrule
Epoch &  \multicolumn{3}{c|}{400} & \multicolumn{2}{c|}{400} & 300 & 1000 & 400 \\ 
Batch Size &  \multicolumn{3}{c|}{4096} & \multicolumn{2}{c|}{256} & 256 &256 / 512  & 256 / 512 \\ 

ImageNet-1K (Linear)        & 69.8 & 73.2 & 70.7 & 71.0 & 70.8 & 70.7 & 68.7 / 71.7 & \cellcolor{pearDark!20 } \textbf{71.1 / 72.3} \\ 
\bottomrule
\end{tabular}
}
\end{table}

\smallskip \noindent \textbf{Generalization of DCL to Different Domains.}
DCL can be easily adapted to different domains (e.g., speech and language models) to achieve competitive performance. We demonstrate that DCL can be combined with SOTA SSL speech models, e.g., wav2vec 2.0~\cite{wav2vec} which uses transformer backbone and requires enormous computation resources. We evaluate wav2vec 2.0 on its downstream tasks and perform better by applying the DCL method. Detailed results and discussion can be found in Appendix. To the best of our knowledge, DCL can be potentially combined with a transformer-based language model, CLIP~\cite{RadfordKHRGASAM21}, which uses a very large batch size of 32768. With DCL, CLIP shall maintain its complexity and achieve huge learning efficiency when the batch size becomes smaller. Note that it has been implemented by~\cite{xclip}.

\smallskip \noindent \textbf{DCL Convergence for Large Batch Sizes.}
The performance of DCL appears to have less gain compared to InfoNCE-based baseline when the batch size is large. According to Figure~\ref{fig:coupling} and the theoretical analysis, the reason is that the NPC multiplier $q_{B} \to 0$ when the batch size is large (e.g., 1024). As shown in the analysis, InfoNCE loss converges to the DCL loss as the batch size approaches infinity. With 400 training epochs, the ImageNet-1K top-1 accuracy slightly increases from 69.5\% to 69.9\% when the batch size increases from 256 to 1024. Please refer to Table~\ref{tab:largebatchsize}.

\begin{table}[t!]
    \centering
    \caption{kNN \& linear top-1 accuracy ($\%$) comparison of SSL approaches on CIFAR10 and ImageNet-100.}
    \label{tab:compare-cifar-img100}
    \resizebox{\textwidth}{!}{
    \begin{tabular}{l|ccccccc}
     \toprule
     ResNet-18 @ 256 Batch Size & DINO & SwAV & SimSiam & VICReg & Barlow Twins & NNCLR & NNCLR+DCL \\
     \midrule
     \midrule
     CIFAR10, 1000 Epoch (kNN) & 89.5 & 89.2 & 90.5 & 92.1 & 92.1 & 91.8 & \cellcolor{pearDark!20}\textbf{92.3} \\
     ImageNet-100, 400 Epoch (Linear) & 74.9 & 74.0 & 74.5 & 79.2 & 80.2 & 79.8 & \cellcolor{pearDark!20}\textbf{80.6} \\
     \bottomrule
    \end{tabular}
    }
\end{table}

\begin{table}[t!]
    \centering
    \caption{Results of DCL and SimCLR with large batch size and learning epochs.}
    \label{tab:largebatchsize}
   
    \begin{tabular}{l|ccc}
    \toprule
        ImageNet-1K (ResNet-50) & Batch Size & Epoch & Top-1 Accuracy (\%) \\
        \midrule
        \midrule
        SimCLR                & 256        & 200    & 61.8 \\
        SimCLR            & 256        & 400    & 64.8 \\ 
        SimCLR          & 1024        & 400    & 67.3 \\
        \midrule
        SimCLR w/ DCL                & 256        & 200    & 67.8 (\textcolor{pearDark}{\bf+6.0}) \\
        SimCLR w/ DCL             & 256        & 400    & 69.5 (\textcolor{pearDark}{\bf+4.7}) \\ 
        SimCLR w/ DCL          & 1024        & 400    & 69.9 (\textcolor{pearDark}{\bf+2.6}) \\
        \bottomrule
    \end{tabular} 
\end{table}

\section{Conclusion}
This paper identifies the negative-positive-coupling (NPC) effect in the widely used InfoNCE loss, making the SSL task significantly easier to solve with smaller batch size. By removing the NPC effect, we reach a new objective function, \textit{decoupled contrastive learning} (DCL). The proposed DCL loss function requires minimal modification to the SimCLR baseline and provides efficient, reliable, and nontrivial performance improvement on various benchmarks. Given the conceptual simplicity of DCL and that it requires neither momentum encoding, large batch size, or long epochs to reach competitive performance. Notably, DCL can be combined with the SOTA contrastive learning method, NNCLR, to achieve $72.3\%$ ImageNet-1K top-1 accuracy with $512$ batch size in $400$ epochs. We wish that DCL can serve as a strong baseline for the contrastive-based SSL methods. Further, an important lesson from the DCL loss is that a more efficient SSL task shall maintain its complexity when the batch size becomes smaller. \\

\noindent{\bf Acknowledgements}. This work was supported in part by the MOST grants 110-2634-F-007-027, 110-2221-E-001-017 and 111-2221-E-001-015 of Taiwan. We are grateful to National Center for High-performance Computing and Meta AI Research for providing computational resources and facilities.

\clearpage
%
%
\bibliographystyle{splncs04}
\bibliography{egbib}

\appendix


\onecolumn
\setcounter{page}{1}
\section{Appendix}
\subsection{Proof of proposition 1}
\label{sub:a.1}

\small
\begin{equation*}
\centering
L_i^{(k)} = -\log{\frac{\exp(\langle \mathbf{z}_i^{(1)},\mathbf{z}_i^{(2)} \rangle/\tau)}{\exp(\langle \mathbf{z}_i^{(1)},\mathbf{z}_i^{(2)} \rangle/\tau) + \sum_{l\in\{1,2\}, j\in[\![1,N]\!],j\neq i}{\exp(\langle \mathbf{z}_i^{(k)},\mathbf{z}_j^{(l)} \rangle/\tau)}}}
\label{eq:simclr_loss_re}
\end{equation*}
\normalsize

{\bf Proposition 1.}
\label{prop:coupling2}
There exists a negative-positive coupling (NPC) multiplier $q_{B,i}^{(1)}$ in the gradient of $L_i^{(1)}$:

\tiny
\begin{align*}
\begin{cases} 
-\nabla_{\mathbf{z}_{i}^{(1)}}L_{i}^{(1)} = \frac{q_{B,i}^{(1)}}{\tau}\left[ \mathbf{z}_i^{(2)} - \sum_{l\in\{1,2\}, j\in[\![1,N]\!],j\neq i}{\frac{\exp{\langle \mathbf{z}_i^{(1)},\mathbf{z}_j^{(l)} \rangle/\tau}}{\sum_{q\in\{1,2\}, j\in[\![1,N]\!], j\neq i}{\exp(\langle \mathbf{z}_i^{(1)},\mathbf{z}_j^{(q)} \rangle/\tau)}}}\cdot \mathbf{z}_j^{(l)}\right]\\
-\nabla_{\mathbf{z}_{i}^{(2)}}L_{i}^{(1)} = \frac{q_{B,i}^{(1)}}{\tau}\cdot \mathbf{z}_i^{(1)}\\
-\nabla_{\mathbf{z}_{j}^{(l)}}L_{i}^{(1)} = - \frac{q_{B,i}^{(1)}}{\tau}\frac{\exp{\langle \mathbf{z}_i^{(1)},\mathbf{z}_j^{(l)} \rangle/\tau}}{\sum_{q\in\{1,2\}, j\in[\![1,N]\!],j\neq i}{\exp(\langle \mathbf{z}_i^{(1)},\mathbf{z}_j^{(q)} \rangle/\tau)}}\cdot \mathbf{z}_i^{(1)}
\end{cases}
\end{align*}
\normalsize
where the NPC multiplier $q_{B,i}^{(1)}$ is:
\begin{align}
q_{B,i}^{(1)} = 1 - \frac{\exp(\langle \mathbf{z}_i^{(1)},\mathbf{z}_i^{(2)} \rangle/\tau)}{\exp(\langle \mathbf{z}_i^{(1)},\mathbf{z}_i^{(2)}\rangle/\tau) + \sum_{q\in\{1,2\}, j\in[\![1,N]\!],j\neq i}{\exp(\langle \mathbf{z}_i^{(1)},\mathbf{z}_j^{(q)} \rangle/\tau)}}
\end{align}
Due to the symmetry, a similar NPC multiplier $q_{B,i}^{(k)}$ exists in the gradient of $L_i^{(k)}, k\in\{1,2\}, i\in[\![1,N]\!]$.
\begin{proof}
$ $\newline
Let
$Y_{i,1} = \exp(\langle \mathbf{z}_i^{(1)},\mathbf{z}_i^{(2)}\rangle/\tau) + \sum_{q\in\{1,2\}, j\in[\![1,N]\!],j\neq i}{\exp(\langle \mathbf{z}_i^{(1)},\mathbf{z}_j^{(q)} \rangle/\tau)}$,  $U_{i,1}=\sum_{q\in\{1,2\}, j\in[\![1,N]\!],j\neq i}{\exp(\langle \mathbf{z}_i^{(1)},\mathbf{z}_j^{(q)} \rangle/\tau)}$. 
So $q_{B,i}^{(1)} = \frac{U_{i,1}}{Y_{i,1}}$.

\tiny
\begin{align*}
    -\nabla_{\mathbf{z}_{i}^{(1)}}L_{i}^{(1)} &= \frac{\mathbf{z}_i^{(2)}}{\tau} - \frac{1}{Y_{i,1}}\cdot\exp(\langle \mathbf{z}_i^{(1)},\mathbf{z}_i^{(2)}\rangle/\tau)\cdot \frac{\mathbf{z}_i^{(2)}}{\tau} - \frac{1}{Y_{i,1}}\cdot \sum_{q\in\{1,2\}, j\in[\![1,N]\!],j\neq i}{\exp(\langle \mathbf{z}_i^{(1)},\mathbf{z}_j^{(q)} \rangle/\tau)}\frac{\mathbf{z}_j^{(q)}}{\tau}\\
    &=(1-\frac{1}{Y_{i,1}}\cdot\exp(\langle \mathbf{z}_i^{(1)},\mathbf{z}_i^{(2)}\rangle/\tau))\frac{\mathbf{z}_{i}^{(2)}}{\tau} - \frac{1}{Y_{i,1}}\cdot \sum_{q\in\{1,2\}, j\in[\![1,N]\!],j\neq i}{\exp(\langle \mathbf{z}_i^{(1)},\mathbf{z}_j^{(q)} \rangle/\tau)}\frac{\mathbf{z}_j^{(q)}}{\tau}\\
    &=\frac{U_{i,1}}{Y_{i,1}}\frac{\mathbf{z}_{i}^{(2)}}{\tau} - \frac{U_{i,1}}{Y_{i,1}}\cdot\frac{1}{U_{i,1}}\cdot \sum_{q\in\{1,2\}, j\in[\![1,N]\!],j\neq i}{\exp(\langle \mathbf{z}_i^{(1)},\mathbf{z}_j^{(q)} \rangle/\tau)}\frac{\mathbf{z}_j^{(q)}}{\tau}\\
    &= \frac{1}{\tau}\frac{U_{i,1}}{Y_{i,1}}\left[\mathbf{z}_i^{(2)} - \sum_{q\in\{1,2\}, j\in[\![1,N]\!],j\neq i}{\frac{\exp(\langle \mathbf{z}_i^{(1)},\mathbf{z}_j^{(q)} \rangle/\tau)}{U_{i,1}} \cdot \mathbf{z}_j^{(q)}}\right]\\
    &= \frac{q_{B,i}^{(1)}}{\tau}\left[\mathbf{z}_i^{(2)} - \sum_{q\in\{1,2\}, j\in[\![1,N]\!],j\neq i}{\frac{\exp(\langle \mathbf{z}_i^{(1)},\mathbf{z}_j^{(q)} \rangle/\tau)}{U_{i,1}} \cdot \mathbf{z}_j^{(q)}}\right]
\end{align*}
\normalsize

\small
\begin{align*}
    -\nabla_{\mathbf{z}_{i}^{(2)}}L_{i}^{(1)} &= \frac{1}{\tau}\mathbf{z}_{i}^{(1)} - \frac{1}{Y_{i,1}}\exp(\langle \mathbf{z}_i^{(1)},\mathbf{z}_i^{(2)}\rangle/\tau)\cdot\frac{\mathbf{z}_i^{(1)}}{\tau}\\
    & = \frac{1}{\tau}\left(1 - \frac{1}{Y_{i,1}}\exp(\langle \mathbf{z}_i^{(1)},\mathbf{z}_i^{(2)}\rangle/\tau)\right)\cdot\mathbf{z}_i^{(1)}\\
    & = \frac{1}{\tau}\frac{U_{i,1}}{Y_{i,1}}\cdot\mathbf{z}_i^{(1)}\\
    &= \frac{q_{B,i}^{(1)}}{\tau}\cdot \mathbf{z}_i^{(1)}
\end{align*}
\normalsize

\small
\begin{align*}
-\nabla_{\mathbf{z}_{j}^{(l)}}L_{i}^{(1)} &= \frac{1}{Y_{i,1}}\exp(\langle \mathbf{z}_i^{(1)},\mathbf{z}_j^{(q)} \rangle/\tau)\cdot \frac{\mathbf{z}_i^{(1)}}{\tau}\\
&= \frac{U_{i,1}}{Y_{i,1}}\cdot\frac{1}{U_{i,1}}\exp(\langle \mathbf{z}_i^{(1)},\mathbf{z}_j^{(q)} \rangle/\tau)\cdot \frac{\mathbf{z}_i^{(1)}}{\tau}\\
&= \frac{q_{B,i}^{(1)}}{\tau}\cdot \frac{\exp(\langle \mathbf{z}_i^{(1)},\mathbf{z}_j^{(q)} \rangle/\tau)}{U_{i,1}} \mathbf{z}_i^{(1)}
\end{align*}
\normalsize
where we can easily see that $\sum_{q\in\{1,2\}, j\in[\![1,N]\!],j\neq i}{\frac{\exp(\langle \mathbf{z}_i^{(1)},\mathbf{z}_j^{(q)} \rangle/\tau)}{U_{i,1}}} = 1$.
\end{proof}

\clearpage

\subsection{Proof of proposition 2}
\label{sub:a.2}
\noindent{\bf Proposition 2.}
Removing the positive pair from the denominator of Equation~3 leads to a decoupled contrastive learning loss. If we remove the NPC multiplier $q_{B,i}^{(k)}$ from Equation~3, we reach a decoupled contrastive learning loss $L_{DC} = \sum_{k\in\{1,2\}, i\in[\![1,N]\!]}{L_{DC,i}^{(k)}}$, where $L_{DC,i}^{(k)}$ is:
\begin{align*}
L_{DC,i}^{(k)} &= -\log{\frac{\exp(\langle \mathbf{z}_i^{(1)},\mathbf{z}_i^{(2)} \rangle/\tau)}{\bcancel{\exp(\langle \mathbf{z}_i^{(1)},\mathbf{z}_i^{(2)} \rangle/\tau)} + U_{i,k}}} \\
               &= -\langle \mathbf{z}_i^{(1)},\mathbf{z}_i^{(2)} \rangle/\tau + \log{U_{i,k}}
\label{eq:DC1}
\end{align*}
where $U_{i,k} = \sum_{l\in\{1,2\}, j\in[\![1,N]\!], j\neq i}{\exp(\langle \mathbf{z}_i^{(k)},\mathbf{z}_j^{(l)} \rangle/\tau)}$.
\begin{proof}
By removing the positive term in the denominator of Equation~1, we can repeat the procedure in the proof of Proposition~1 and see that the coupling term disappears.
\end{proof}

\begin{table}[t!]
\centering
\caption{\label{tab:imgnet_r50} ImageNet-1K top-1 accuracies ($\%$) of linear classifiers trained on representations of different SSL methods with ResNet-50 backbone. The results in the lower section are the same methods with a large-scale experiment setting. We find that given lower computational budget, DCL model are better than other state-of-the-arts approaches. Its effectiveness \textbf{does not rely on} large batch size and learning epochs (SimCLR~\cite{chen2020simple}, NNCLR~\cite{dwibedi2021little}), momentum encoding (BYOL~\cite{GrillSATRBDPGAP20}, MoCo-v2~\cite{chen2020improved}), or other tricks such as stop-gradient (SimSiam~\cite{abs-2011-10566}) and multi-cropping (SwAV~\cite{caron2020unsupervised}).}
\resizebox{0.8\textwidth}{!}{
\begin{tabular}{l|ccccc}
\toprule
Method &
Param. (M) &
Batch Size &
Epochs &
Top-1 Linear ($\%$)
\\
\hline
NPID~\cite{wu2018unsupervised}
& 24 & 256 & 200 & 56.5 \\
MoCo~\cite{he2020momentum} 
& 24 & 256 & 200 & 60.6 \\
CMC~\cite{tian2019contrastive}
& 47 & 256 & 280 & 64.1 \\
MoCo-v2~\cite{chen2020improved}
& 28 & 256 & 200 & 67.5 \\
SwAV~\cite{caron2020unsupervised}
& 28 & 4096 & 200 & 69.1 \\
SimSiam~\cite{abs-2011-10566}
& 28 & 256 & 200 & 70.0 \\
InfoMin~\cite{tian2020makes}
& 28 & 256 & 200 & 70.1 \\
BYOL~\cite{GrillSATRBDPGAP20}
& 28 & 4096 & 200 & 70.6 \\
SiMo~\cite{zhu2020eqco}
& 28 & 256 & 200 & 68.0 \\
Hypersphere~\cite{wang2020understanding}
& 28 & 256 & 200 & 67.7 \\
SimCLR~\cite{chen2020simple}
& 28 & 256 & 200 & 61.8 \\

SimCLR+DCL           
& \cellcolor{pearDark!20}28 & \cellcolor{pearDark!20}256 & \cellcolor{pearDark!20}200 & \cellcolor{pearDark!20}67.8 \\
SimCLR+DCL(w/ BYOL aug.)           
& \cellcolor{pearDark!20}28 & \cellcolor{pearDark!20}256 & \cellcolor{pearDark!20}200 & \cellcolor{pearDark!20}68.2 \\

\midrule
\midrule

PIRL~\cite{misra2020self}
& 24 & 256 & 800 & 63.6 \\
BYOL~\cite{GrillSATRBDPGAP20}
& 28 & 4096 & 400 & 73.2 \\
SwAV~\cite{caron2020unsupervised}
& 28 & 4096 & 400 & 70.7 \\
MoCo-v2~\cite{chen2020improved}
& 28 & 256 & 400 & 71.0 \\
SimSiam~\cite{abs-2011-10566}
& 28 & 256 & 400 & 70.8 \\
Barlow Twins~\cite{zbontar2021barlow}
& 28 & 256 & 300 & 70.7 \\
SimCLR~\cite{chen2020simple}
& 28 & 4096 & 1000 & 69.3 \\
SimCLR+DCL
&\cellcolor{pearDark!20}28 & \cellcolor{pearDark!20}256 & \cellcolor{pearDark!20}400 & \cellcolor{pearDark!20}69.5 \\
NNCLR~\cite{dwibedi2021little}
& 28 & 256 & 1000 & 68.7 \\
NNLCR+DCL
&\cellcolor{pearDark!20}28 & \cellcolor{pearDark!20}256 & \cellcolor{pearDark!20}400 & \cellcolor{pearDark!20}71.1 \\
NNCLR~\cite{dwibedi2021little}
& 28 & 512 & 1000 & 71.7 \\
NNCLR+DCL
&\cellcolor{pearDark!20}28 & \cellcolor{pearDark!20}512 & \cellcolor{pearDark!20}400 & \cellcolor{pearDark!20}72.3 \\

\bottomrule
\end{tabular}
}

\end{table}

\subsection{Linear classification on ImageNet-1K}
Top-1 accuracies of linear evaluation in Table~\ref{tab:imgnet_r50} shows that, we compare with the state-of-the-art SSL approaches on ImageNet-1K. For fairness, we list each approach's batch size and learning epoch, shown in the original paper. During pre-training, DCL is based on a ResNet-50 backbone, with two views with the size 224 $\times$ 224. DCL relies on its simplicity to reach competitive performance without relatively huge batch sizes and epochs or other pre-training schemes, i.e., momentum encoder, clustering, and prediction head. We report 400-epoch versions of DCL combined with NNCLR~\cite{dwibedi2021little}. It achieves $71.1\%$ under the batch size of 256 and 400-epoch pre-training, which is better than NNCLR~\cite{dwibedi2021little} in their optimal case, $68.7\%$ with a batch size of 256 and 1000-epoch. Note that SwAV~\cite{caron2018deep}, BYOL~\cite{GrillSATRBDPGAP20}, SimCLR, and PIRL~\cite{misra2020self} need a huge batch size of 4096, and SwAV further applies multi-cropping extra views to reach optimal performance. The results of SwAV are taken from SimSiam that multi-cropping is not included.

\subsection{Implementation details}
\label{sub:a.4}
\paragraph{Default DCL augmentations.}
We follow the settings of SimCLR to set up the data augmentations. We use $RandomResizedCrop$ with scale in [0.08, 1.0] and follow by $RandomHorizontalFlip$. Then, $Color Jittering$ with strength in [0.8, 0.8, 0.8, 0.2] with probability of 0.8, and $RandomGrayscale$ with probability of 0.2. $Gaussian Blur$ includes Gaussian kernel with standard deviation in [0.1, 2.0]. 
\paragraph{Strong DCL augmentations.} We follow the asymmetric image augmentation of BYOL to replace default DCL augmentation in ablations. Table~3 demonstrates that the ImageNet-1K top-1 performance is increased from 67.8\% to 68.2\% by applying asymmetric augmentations.

\paragraph{Linear evaluation.}
Following the OpenSelfSup benchmark~\cite{openselfsup}, we first train the linear classifier with batch size 256 for 100 epochs. We use the SGD optimizer with momentum = 0.9, and weight decay = 0. The base $lr$ is set to 30.0 and decay by 0.1 at epoch [60, 80]. We further demonstrate the linear evaluation protocol of SimSiam~\cite{abs-2011-10566}, which raises the batch size to 4096 for 90 epochs. The optimizer is switched to LARS optimizer with base $lr = 1.2$ and cosine decay schedule. The momentum and weight decay have remained unchanged. We found the second one slightly improves the performance.

\subsection{Relation to alignment and uniformity}
\label{sec:align_uniform}
In this section, we provide a thorough discussion of the connection and difference between DCL and Hypersphere~\cite{wang2020understanding}, which does not have negative-positive coupling either. However, there is a critical difference between DCL and Hypersphere, and the difference is that the order of the expectation and exponential is swapped. Let us assume the latent embedding vectors $z$ are normalized for analytical convenience. When $z_{i},z_{j}$ are normalized, $\exp(\langle \mathbf{z}_i^{(k)},\mathbf{z}_i^{(l)} \rangle/\tau)$ and $\exp(-|| \mathbf{z}_i^{(k)}-\mathbf{z}_i^{(l)} ||^2 /\tau)$ are the same, except for a trivial scale difference. Thus we can write $L_{DCL}$ and $L_{align-uni}$ in a similar fashion:

\begin{align*}
L_{DCL} &= L_{DCL,pos} + L_{DCL,neg}
\end{align*}
\begin{align*}
L_{align-uni} = L_{align} + L_{uniform}
\end{align*}
where 
\begin{align*}
\left\{\begin{array}{c}
    L_{DCL,neg}= \sum_{i}\log(\sum_{j\neq i} \exp(\langle \mathbf{z}_i^{(k)},\mathbf{z}_j^{(l)} \rangle/\tau)),   \\
    L_{uniform} = \log(\sum_{i}\sum_{j\neq i} \exp(\langle \mathbf{z}_i^{(k)},\mathbf{z}_j^{(l)} \rangle/\tau)).      
\end{array} \right.
\end{align*}

With the right weight factor, $L_{align}$ can be made exactly the same as $L_{DCL,pos}$. So let’s focus on $L_{DCL,neg}$ and $L_{uniform}$:

\begin{align*}
L_{DCL,neg} = \sum_{i}\log(\sum_{j\neq i} \exp(\langle \mathbf{z}_i^{(k)},\mathbf{z}_j^{(l)} \rangle/\tau))
\end{align*}

\begin{align*}
L_{uniform} = \log(\sum_{i}\sum_{j\neq i} \exp(\langle \mathbf{z}_i^{(k)},\mathbf{z}_j^{(l)} \rangle/\tau))
\end{align*}

Similar to the earlier analysis in the manuscript, the latter $L_{uniform}$ introduces a negative-negative coupling between the negative samples of different positive samples. If two negative samples of $z_{i}$ are close to each other, the gradient for $z_{i}$ would also be attenuated. This behaves similarly to the negative-positive coupling. That being said, while Hypersphere does not have a negative-positive coupling, it has a similarly problematic negative-negative coupling.

A case can simply demonstrate the negative-negative coupling in~\cite{wang2020understanding}. Let's assume the model has the batch size of 3, and temperature $\tau$ is 1. Both $L_{DCL,neg}$ and $L_{uniform}$ can be formulated as follows:

\begin{align*}
L_{DCL,neg} = \log( \exp(\langle \mathbf{z}_1^{(k)},\mathbf{z}_2^{(l)} \rangle) + \exp(\langle \mathbf{z}_1^{(k)},\mathbf{z}_3^{(l)} \rangle)) + \\ \log( \exp(\langle \mathbf{z}_2^{(k)},\mathbf{z}_1^{(l)} \rangle) + \exp(\langle \mathbf{z}_2^{(k)},\mathbf{z}_3^{(l)} \rangle)) + \\ \log( \exp(\langle \mathbf{z}_3^{(k)},\mathbf{z}_1^{(l)} \rangle) + \exp(\langle \mathbf{z}_3^{(k)},\mathbf{z}_2^{(l)} \rangle))
\end{align*}

\begin{align*}
L_{uniform} = \log( \exp(\langle \mathbf{z}_1^{(k)},\mathbf{z}_2^{(l)} \rangle) + \exp(\langle \mathbf{z}_1^{(k)},\mathbf{z}_3^{(l)} \rangle) + \exp(\langle \mathbf{z}_2^{(k)},\mathbf{z}_1^{(l)} \rangle) + \\ \exp(\langle \mathbf{z}_2^{(k)},\mathbf{z}_3^{(l)} \rangle) + \exp(\langle \mathbf{z}_3^{(k)},\mathbf{z}_1^{(l)} \rangle) + \exp(\langle \mathbf{z}_3^{(k)},\mathbf{z}_2^{(l)} \rangle)) 
\end{align*}

If the value of $\exp(\langle \mathbf{z}_1^{(k)},\mathbf{z}_3^{(l)} \rangle)$ is much larger (e.g., hard negatives) than other terms, there would be a huge difference between $L_{DCL,neg}$ and $L_{uniform}$. Since $L_{uniform}$ first sums up all the negative pair samples in the batch together, it may cause the loss to be dominated by a specific negative pair sample. Thus, in the DCL loss, the negative samples from different positives are not coupled in contrast to the uniformity loss in~\cite{wang2020understanding}. 

Next, we provide a comprehensive empirical comparison. The empirical experiments match the analytical prediction: DCL outperforms Hypersphere with a more considerable margin under a smaller batch size. 

The comparisons of DCL to Hypersphere are evaluated on STL10, ImageNet-100, ImageNet-1K under various settings. For STL10 data, we implement DCL based on the official code of Hypersphere. The encoder and the hyperparameters are the same as Hypersphere, which has not been optimized for DCL in any way. We have found that Hypersphere did a pretty thorough hyperparameter search. We believe the default hyperparameters are relatively optimized for Hypersphere.

In Table~\ref{tab:hyper-stl10}, DCL reaches 84.4\% (fc7+Linear) compared to 83.2\% (fc7+Linear) reported in Hypersphere on STL10. In Table~\ref{tab:hyper-moco} and Table~\ref{tab:hyper-mocov2}, DCL achieves better performance than Hypersphere under the same setting (MoCo \& MoCo-v2) on ImageNet-100 data. DCL further shows strong results compared against Hypersphere on ImageNet-1K in Table~\ref{tab:hyper-imgnet1k}. We also provide the STL10 comparisons of DCL and Hypersphere under different batch sizes in Table~\ref{tab:hyper-stl10-batch}. The experiment shows the advantage of DCL becomes larger with smaller batch size. Please note that we did not tune the parameters for DCL at all. This should be a more than fair comparison.

\begin{table}[t!]
    \centering
        \caption{STL10 comparisons Hypersphere and DCL under the same experiment setting.}
    \label{tab:hyper-stl10}
    \begin{tabular}{lcccc}
    \toprule
         STL10 & fc7+Linear & fc7+5-NN &Output + Linear & Output + 5-NN   \\
         \midrule
         \midrule
         Hypersphere & 83.2 & 76.2 &	80.1 & 79.2  \\
         DCL    & \cellcolor{pearDark!20} \bf 84.4 (+1.2) &	\cellcolor{pearDark!20} \bf 77.3 (+1.1) & \cellcolor{pearDark!20} \bf 81.5 (+1.4) &	\cellcolor{pearDark!20} \bf 80.5 (+1.3)  \\
    \bottomrule
    \end{tabular}

\end{table}

\begin{table}[t!]
    \centering
        \caption{ImageNet-100 comparisons of Hypersphere and DCL under the same setting (MoCo).}
    \label{tab:hyper-moco}
    \begin{tabular}{lccc}
    \toprule
         ImageNet-100	&Epoch	&Memory Queue Size	&Linear Top-1 Accuracy (\%)   \\
         \midrule
         \midrule
         Hypersphere & 240 &	16384 & 75.6  \\
         DCL     & \cellcolor{pearDark!20}240 & \cellcolor{pearDark!20}16384 & \cellcolor{pearDark!20}\bf 76.8 (+1.2) \\
    \bottomrule
    \end{tabular}

\end{table}

\begin{table}[t!]
    \centering
        \caption{ImageNet-100 comparisons of Hypersphere and DCL under the same setting (MoCo-v2) except for memory queue size.}
    \label{tab:hyper-mocov2}
    \begin{tabular}{lccc}
    \toprule
         ImageNet-100	&Epoch	&Memory Queue Size	&Linear Top-1 Accuracy (\%)   \\
         \midrule
         \midrule
         Hypersphere & 200 &	16384 & 77.7  \\
         DCL     & \cellcolor{pearDark!20}200 &	\cellcolor{pearDark!20}8192 &	\cellcolor{pearDark!20}\bf 80.5 (+2.7) \\
    \bottomrule
    \end{tabular}

\end{table}

In every single one of the experiments, DCL outperforms Hypersphere. Although the difference between the DCL and Hypersphere is slight, it makes DCL more easier to alleviate the domination from a specific negative pair in a batch. We hope these results show the unique value of DCL compared to Hypersphere.

\begin{table}[t!]
    \centering
        \caption{ImageNet-1K comparisons of and DCL under the best setting. In this experiment both of the methods used their optimized hyperparameters.}
    \label{tab:hyper-imgnet1k}
    \begin{tabular}{lccc}
    \toprule
         ImageNet-1K	&Epoch	&Batch Size	&Linear Top-1 Accuracy (\%)   \\
         \midrule
         \midrule
         MoCo-v2 Baseline & 200 &	256 (Memory queue = 65536)	  & 67.5 \\
         Hypersphere & 200 &	256 (Memory queue = 65536)	  & 67.7 (+0.2) \\
         DCL     & \cellcolor{pearDark!20}200 &	\cellcolor{pearDark!20}256  & \cellcolor{pearDark!20}\bf 68.2 (+0.7) \\
    \bottomrule
    \end{tabular}

\end{table}

\begin{table}[t!]
    \centering
        \caption{STL10 comparisons of Hypersphere and DCL under different batch sizes.}
    \label{tab:hyper-stl10-batch}
    \begin{tabular}{lccccc}
    \toprule
         Batch Size & 32 & 64 & 128 & 256 & 768   \\
         \midrule
         \midrule
         Hypersphere & 78.9&	81.0 &	81.9&	82.6&	83.2  \\
         DCL     & \cellcolor{pearDark!20}\bf 81.0 (+2.1) & \cellcolor{pearDark!20}\bf 82.9 (+1.9) & \cellcolor{pearDark!20}\bf 83.7 (+1.8) & \cellcolor{pearDark!20}\bf 84.2 (+1.6) & \cellcolor{pearDark!20}\bf 84.4 (+1.2)
  \\
    \bottomrule
    \end{tabular}

\end{table}

\begin{table}[t!]
    \centering
    \caption{Results of DCL on wav2vec 2.0 be evaluated on two downstream tasks.}
\label{tab:wav2vec}
    \begin{threeparttable}
    \begin{tabular}{l|cc}
    \toprule
        Downstream task (Accuracy) & Speaker Identification$^\dag$ (\%) & Intent Classification$^\ddag$ (\%)   \\
        \midrule
        \midrule
        wav2vec 2.0 Base Baseline & 74.9 & 92.3 \\
        wav2vec 2.0 Base w/ (DCL)              & \cellcolor{pearDark!20}{\textbf{75.2}} & \cellcolor{pearDark!20}{\textbf{92.5}} \\

        \bottomrule
    \end{tabular}
    \begin{tablenotes}
    \item[\dag] In the downstream training process, the pre-trained representation first mean-pool and forward a fully a connected layer with cross-entropy loss on the VoxCeleb1~\cite{NagraniCXZ20}.
    \item[\ddag] In the downstream training process, the pre-trained representation first mean-pool and forward a fully a connected layer with cross-entropy loss on Fluent Speech Commands~\cite{LugoschRITB19}.
    \end{tablenotes}
\end{threeparttable}

\end{table}

\begin{table}[t!]
    \centering
        \caption{Comparisons between the cross entropy and DCL in supervised classifier under different numbers of batch sizes (32, 128, and 256).}
    \label{tab:superivsedcifar}
    \begin{tabular}{c|ccc|ccc}
    \toprule
         Architecture@epoch &  \multicolumn{6}{c}{ResNet-20@200 epoch} \\
        \midrule
        Batch Size &  32  &  \quad 128 \quad  & \quad 256 \quad & 32  &  \quad 128 \quad  & 256 \\
    \midrule
    \midrule
        Dataset & \multicolumn{3}{c}{CIFAR10 (top-1)} & \multicolumn{3}{|c}{CIFAR100 (top-1)}\\
        \midrule
        Cross entropy   &  91.5   &  92.3  & 91.0 & 61.9  & 62.7 & 61.8 \\
        DCL             & \cellcolor{pearDark!20}{89.2}  & \cellcolor{pearDark!20}{91.4} & \cellcolor{pearDark!20}{91.2}  & \cellcolor{pearDark!20}{60.2}  & \cellcolor{pearDark!20}{61.8} & \cellcolor{pearDark!20}{61.4}\\
    \bottomrule
    \end{tabular}

\end{table}

\subsection{DCL on speech models}

The SOTA SSL speech models, e.g., wav2vec 2.0~\cite{wav2vec} still uses contrastive loss in the objective function. In Table~\ref{tab:wav2vec}, we show the effectiveness of DCL with wav2vec 2.0~\cite{wav2vec}. We replace the InfoNCE loss with the DCL loss and train a wav2vec 2.0 base model (i.e., 7-Conv + 24-Transformer) from scratch.\footnote{The experiment is downscaled to 8 V100 GPUs rather than 64.} After the pre-training of model, we evaluate the representation on two downstream tasks, speaker identification and intent classification. Table~\ref{tab:wav2vec} shows the representation improvement of DCL.

\subsection{Supervised classifier: DCL vs Cross-Entropy}
The idea of DCL, removing positive from the denominator, can also be applied for learning objective function in the supervised classifier. By following~\cite{Idelbayev18a}, we implement the proposed DCL idea on cross entropy loss by removing the positive logits from the denominator of the softmax function. In Table~\ref{tab:superivsedcifar}, it is observed that our supervised DCL achieves slightly lower performance while comparing to the cross-entropy on CIFAR data. One possible reason for undermined performance of DCL in supervised learning might be the different feature interaction between supervised and unsupervised classifiers, which are referred to as parametric and non-parametric classifiers in~\cite{wu2018unsupervised}. 

Under the parametric formulation in~\cite{wu2018unsupervised}, the logits equal to $w^{T}z$, where $w$ is a weight vector for each class and $z$ is the output embedding of the neural network. While in contrastive learning (i.e., non-parametric classifier), the logits equal to $z^{(1)}z^{(2)}$, where $z^{(1)}$ and $z^{(2)}$ are two augmented views of the same sample. In the embedding space of the early training stage, $w$ is relatively far away from $z$ compared to the relation between $z^{(1)}$ and $z^{(2)}$. Consider the effect of NPC multiplier $q_{b}$ into parametric and non-parametric classifier, $q_{b}\rightarrow1$ in parametric classifier might diminish the effectiveness of DCL idea as the coupling effect is already tiny.

\subsection{Ablations of DCLW}
Based on weighting function for the positive pairs in the Section~3 of the manuscript, we provide an another weighting function of DCLW:
\begin{align}
L_{DCLW} = \sum_{k\in\{1,2\}, i\in[\![1,N]\!]}{L_{DCLW,i}^{(i,k)}}
\end{align}    

\begin{align}
L_{DCLW,i}^{(k)} = -w(\mathbf{z}_i^{(1)},\mathbf{z}_i^{(2)})(\langle \mathbf{z}_i^{(1)},\mathbf{z}_i^{(2)} \rangle/\tau) + \log{U_{i,k}}
\end{align}
where $w(\mathbf{z}_i^{(1)},\mathbf{z}_i^{(2)}) = \delta\cdot\exp(-\sigma\cdot\langle \mathbf{z}_i^{(1)},\mathbf{z}_i^{(2)}\rangle )$. Basically, the goal is similar to DCLW that we provide larger weight to hard positives (e.g., a positive pair of samples are far away from each other). 

The results indicate that $\delta=3$ and $\sigma=0.5$ can achieve $85.4\%$ kNN top-1 accuracy on CIFAR10, and outperform the InfoNCE baseline (SimCLR) by $4\%$.

\subsection{Additional Discussion}

\begin{table}[t]
\centering
\caption{The ablation study of various temperatures $\tau$ on CIFAR10.}
\label{tab:ablation_tau}
\begin{tabular}{l|cccccccccccc}
\toprule
     Temperature $\tau$ & 0.07 & 0.1 & 0.2 & 0.3 & 0.4 & 0.5 & 0.6 & 0.7 & 0.8 & 0.9 & 1.0 & Std  \\
     \midrule
     \midrule
     SimCLR & 83.6 & 87.5 & 89.5 & 89.2 & 88.7 & 89.1 & 88.5 & 87.6 & 86.8 & 85.9 & 85.3 & 1.44  \\
     SimCLR w/ DCL    & \cellcolor{pearDark!20}{\bf 88.3
     }& \cellcolor{pearDark!20}{\bf 89.4} & \cellcolor{pearDark!20}{\bf 90.8} & \cellcolor{pearDark!20}{\bf 89.9} & \cellcolor{pearDark!20}{\bf 89.6} & \cellcolor{pearDark!20}{\bf 90.3} & \cellcolor{pearDark!20}{\bf 89.6} & \cellcolor{pearDark!20}{\bf 89.0} & \cellcolor{pearDark!20}{\bf 88.5} & \cellcolor{pearDark!20}{\bf 88.0} & \cellcolor{pearDark!20}{\bf 87.7} & \cellcolor{pearDark!20}{\bf{0.98}}  \\
\bottomrule
\end{tabular}
\end{table}

\smallskip \noindent \textbf{Analysis of Temperature $\tau$.}
In Table~\ref{tab:ablation_tau}, we further provide extensive analysis on temperature $\tau$ in the objective function to support that the DCL method is not sensitive to hyperparameters compared against the InfoNCE-based baselines. In the following, show the temperature $\tau$ search on both DCL and SimCLR on CIFAR10 data. Specifically, we pre-train the network with temperature $\tau$ in $\left \{ 0.07, 0.1, 0.2, 0.3, 0.4, 0.5, 0.6, 0.7, 0.8, 0.9, 1.0 \right \}$ and report results with kNN eval, batch size 512, and 500 epochs. As shown in Table~\ref{tab:ablation_tau}, compared to SimCLR, DCL is less sensitive to hyperparameters, e.g., temperature $\tau$.

\begin{figure}[t]
    \centering
    \includegraphics[width=\textwidth]{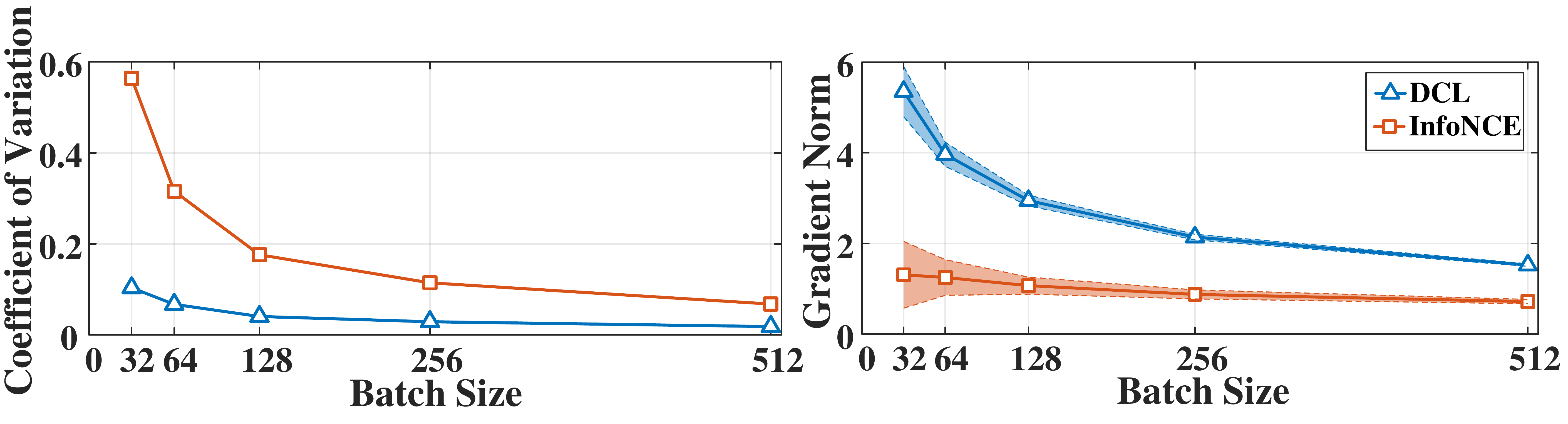}
    \caption{(a) The coefficient of variation ($C_v = \sigma / \mu$) of gradient and (b) the mean gradient norm with its std of baseline (InfoNCE) and proposed method (DCL) under different batch sizes.}
    \label{fig:grident}
\end{figure}

\smallskip \noindent \textbf{Analysis of Gradient.}
For further analysis of the phenomenon of DCL, we visualize the mean norm with its std of the last convolutional layers from the last two residual blocks of ResNet-18 trained on CIFAR100 under different batch sizes. The results in Figure~\ref{fig:grident} show that DCL constantly achieves larger gradients than baseline (InfoNCE) loss, especially under small batch sizes.

\end{document}